\documentclass{article}

    \usepackage[final]{neurips_2025}

\usepackage{fancybox}
\usepackage[utf8]{inputenc} 
\usepackage[T1]{fontenc}    
\PassOptionsToPackage{hyphens}{url}\usepackage{hyperref} 
\usepackage{booktabs}       
\usepackage{amsfonts}       
\usepackage{nicefrac}       
\usepackage{microtype}      
\usepackage{xcolor}         
\usepackage{mathrsfs}
\usepackage{amssymb}
\usepackage{todonotes}
\usepackage{amsmath}

\usepackage{graphicx}
\usepackage{subfigure}
\usepackage{makecell}
\usepackage{algorithm,algorithmic}
\usepackage{paralist,amsmath, amssymb,bm}
\usepackage{multirow}
\usepackage{ntheorem}

\newtheorem{prop}{Proposition}

\newtheorem*{proof}{Proof}

\usepackage{enumitem}
\usepackage{textcase}
\usepackage{booktabs}
\usepackage{fancybox}
\usepackage{wrapfig,lipsum}
\usepackage[normalem]{ulem}
\usepackage[table]{xcolor} 

\usepackage{amsopn}
\usepackage{subcaption}

\usepackage[compact]{titlesec}  
\titlespacing*{\section} {0pt}{0pt}{0pt}
\titlespacing*{\subsection}{0pt}{0pt}{0pt}
\titlespacing*{\subsubsection}{0pt}{0pt}{0pt}

\def\cB{\mathcal{B}}
\def\cD{\mathcal{D}}
\def\cJ{\mathcal{J}}

\def \po{\pi_{\text{old}}}

\def\E{\mathbb E}
\def\R{\mathbb R}

\def\D{\mathbb D}

\def\I{\mathbb I}

\def\p{\mathbf{p}}

\title{DisCO: Reinforcing Large Reasoning Models with Discriminative Constrained Optimization}

\author{%
  Gang Li$^1$\quad\quad Ming Lin\quad\quad  Tomer Galanti$^1$\quad\quad  Zhengzhong Tu$^1$\quad\quad  Tianbao Yang$^1$ \\
  $^1$ Texas A\&M University\\
  \texttt{\{gang-li, galanti, tzz, tianbao-yang\}@tamu.edu} \\
  \texttt{linming04@gmail.com}
}

\begin{document}

\maketitle

\begin{abstract}

The recent success and openness of DeepSeek-R1 have brought widespread attention to Group Relative Policy Optimization (GRPO) as a reinforcement learning method for large reasoning models (LRMs). In this work, we analyze the GRPO objective under a binary reward setting and reveal an inherent limitation of question-level difficulty bias arising from its group relative advantage function. We also identify a connection between GRPO and traditional discriminative methods in supervised learning. Motivated by these insights, we introduce a new \textbf{\underline{Dis}criminative \underline{C}onstrained \underline{O}ptimization (DisCO)} framework for reinforcing LRMs, grounded in the principle of discriminative learning: increasing the scores of positive answers while decreasing those of negative ones. The main differences between DisCO and GRPO and its recent variants are: (1) it replaces the group relative objective with a discriminative objective defined by a scoring function; (2) it abandons clipping-based surrogates in favor of non-clipping RL surrogate objectives used as scoring functions; (3) it employs a simple yet effective constrained optimization approach to enforce the KL divergence constraint.  As a result, DisCO offers notable advantages over GRPO and its variants: (i)  it completely eliminates difficulty bias by adopting discriminative objectives; (ii) it addresses the entropy instability in GRPO and its variants through the use of non-clipping scoring functions and a constrained optimization approach, yielding long and stable training dynamics; (iii) it allows the incorporation of advanced discriminative learning techniques to address data imbalance, where a significant number of questions have more negative than positive generated answers during training. Our experiments on enhancing the mathematical reasoning capabilities of SFT-finetuned models show that DisCO significantly outperforms GRPO and its improved variants such as DAPO, achieving average gains of 7\% over GRPO and 6\% over DAPO across six benchmark tasks for a 1.5B model.\footnote{The code is available at: {\url{https://github.com/Optimization-AI/DisCO} } }

\end{abstract}
\setlength{\abovedisplayskip}{4pt}
\setlength{\belowdisplayskip}{4pt}
\setlength{\belowdisplayshortskip}{4pt}
\section{Introduction}
The recent success and openness of DeepSeek-R1 have sparked a surge of interest in large reasoning models (LRMs), particularly in the context of fine-tuning via reinforcement learning (RL)~\cite{guo2025deepseek}. The core approach involves iteratively generating synthetic data using the reasoning model and applying a rule-based reward mechanism to label the outputs. These rewards are then used to update the policy, \textit{i.e.}, the reasoning model itself. Notably, this framework, featuring a novel policy optimization method called Group Relative Policy Optimization (GRPO),  has enabled DeepSeek-R1 to achieve performance comparable to advanced proprietary LRMs at that time such as OpenAI-o1 on many reasoning benchmarks. 
As a result, GRPO has rapidly become a focal point for advancing LRM capabilities, particularly in domains like mathematics and scientific reasoning.

Several efforts have sought to replicate the performance of DeepSeek-R1 or to further enhance reasoning models using GRPO~\cite{wen2025light, openr1, deepscaler2025,skywork-or1-2025, xiaomi2025mimo,xue2025dancegrpounleashinggrpovisual,bercovich2025llamanemotronefficientreasoningmodels}, while few others have tried to identify its inherent limitations with potential remedies~\cite{liu2025understanding,yu2025dapo, lin2025cppo}. Useful tricks have been introduced to improve GRPO~\cite{liu2025understanding, deepcoder2025,yu2025dapo,skywork-or1-2025, chu2025gpg, zhang2025grpo}. For instance, DAPO~\cite{yu2025dapo} employs two distinct clipping hyperparameters to mitigate {\it entropy collapse}, encouraging exploration. Dr. GRPO~\cite{liu2025understanding} removes the variance normalization in advantage function, aiming to mitigate the issue of {\it difficulty bias}.  However, these approaches remain heuristic and ad-hoc, lacking a principled foundation and falling short of fully addressing GRPO’s inherent limitations. Our analysis identifies that Dr. GRPO continues to suffer from the difficulty bias issue, while our experiments show that DAPO may induce excessive entropy growth, producing highly random outputs. This motivates us to explore a central question: 

\doublebox{
  \begin{minipage}{0.96\textwidth}
\it How can we design more effective optimization methods for reinforcing large reasoning models in a principled manner without inheriting the limitations of GRPO?
\end{minipage}}

This paper addresses the above question through a complete redesign of the objective function, grounded in the principles of discriminative learning. Specifically, we first analyze the objective function of GRPO and its variants under {\bf a binary reward setting}, leading to two key insights: \textbf{(1)} the root cause of GRPO's difficulty bias lies in its group relative advantage function, which induces disproportionately small weights to questions that are either too easy or too hard; and \textbf{(2)} there exists a conceptual connection to traditional discriminative approaches in AUC maximization, which aim to increase the scores of positive outputs while decreasing that of negative outputs. 

Building upon these insights, we propose a principled optimization framework for reinforcing large reasoning models based on discriminative learning. Specifically, we optimize a discriminative objective using a proper scoring function over input-output pairs, which increases the score of positive outputs and decreases that of negative ones. The flexibility of our framework allows us to leverage simple non-clipping RL surrogate objectives as scoring functions without suffering from entropy instability, and to incorporate advanced discriminative techniques to address data imbalance in generated rollouts. To ensure training stability, we adopt a simple yet effective constrained optimization method to enforce a trust region constraint bounding the KL divergence between the updated model and the old model.  Our experiments for mathematical reasoning show that DisCO significantly outperforms all baselines for fine-tuning DeepSeek-R1-Distill-Qwen and -Llama models with a maximum 8k response length for both training and inference, and also achieves a better performance than GRPO that uses a maximum 24k length for training and 32k length for inference.  

Our main contributions are summarized as follows:
\begin{itemize}[leftmargin=*, itemsep=-0.6pt]
\vspace*{-0.1in}
\item We present an \textbf{analysis of GRPO’s objective function}, identifying the root cause of difficulty bias and revealing its conceptual connection to classic discriminative methods for AUC maximization.
\item  We introduce a \textbf{principled discriminative constrained optimization framework} for reinforcing large reasoning models, which avoids both difficulty bias and training instability. This framework gives rise to a family of methods we refer to as {\bf DisCO}.
\item  We demonstrate \textbf{significant improvements} of our DisCO method over GRPO and four other baselines, including DAPO, through experiments for fine-tuning LRMs on mathematical reasoning tasks, with evaluations across six benchmarks.  
\end{itemize}


\section{Related Work}

\textbf{Large Reasoning Models (LRMs).} 
Recent advances of LRMs, such as OpenAI o1~\cite{openo1}, DeepSeek-R1~\cite{guo2025deepseek} and Kimi K1.5~\cite{team2025kimi}, have demonstrated strong reasoning capability in solving complex tasks.
Departing from earlier approaches in LLMs, such as Chain-of-thought (CoT) prompting~\cite{wei2022chain,muennighoff2025s1, zelikman2022star}, Tree-of-Thought~\cite{yao2023tree}, Monte Carlo Tree Search~\cite{feng2023alphazero, trinh2024solving, xin2024deepseek}, a major breakthrough was achieved by scaling RL training using verifiable rewards to incentivize LLMs to learn through self-exploration~\cite{guo2025deepseek, team2025kimi}. Inspired by DeepSeek-R1's core algorithm GRPO~\cite{shao2024deepseekmath}, the research community has actively pursued improved techniques for large-scale RL training, focusing primarily on three directions: algorithm design~\cite{yu2025dapo, liu2025understanding, chu2025gpg, lin2025cppo, team2025kimi, su2025trust}, reward curation~\cite{zhang2025grpo, wen2025light, yu2025dapo}, and sampling strategies~\cite{yu2025dapo, skywork-or1-2025, zhang2025srpo, hu2025open}. Our work falls under the category of algorithm design. 

Among these, Dr. GRPO~\cite{liu2025understanding} identifies response-level length bias and question-level difficulty bias in GRPO algorithm, advocating the removal of length and advantage normalization to improve token efficiency. DAPO~\cite{yu2025dapo} highlights several limitations of GRPO, such as entropy collapse, training instability, and biased loss, and addresses them through techniques like decoupled clipping, dynamic sampling, and a token-level policy loss. GPG~\cite{chu2025gpg} introduces a simplified REINFORCE-based objective that eliminates the need for both the critic and reference models, thereby enhancing scalability for RL training. 
TRPA~\cite{su2025trust} simply uses the Direct Preference Optimization (DPO) objective and a KL divergence regularization for fine-tuning LRMs. It can be recovered from our basic approach, which uses a logistic function as the surrogate loss, the log of likelihood ratio with respect to a frozen reference model as the scoring function, and the KL divergence as a regularization rather than a constraint. However, it does not  address the imbalanced rollouts. The uniqueness and significance of our contributions lie in the analysis of GRPO objective and its variants that reveal key limitations, and the integration of advanced discriminative learning approaches for handling imbalanced rollouts and efficient constrained optimization technique for ensuring training stability.


\textbf{Reinforcement Learning (RL).}  RL is a learning paradigm centered on control and decision-making, in which an agent optimizes a target objective through trial-and-error interactions with its environment~\cite{cao2024survey}. RL approaches are typically categorized into model-based~\cite{silver2017mastering,racaniere2017imagination,nagabandi2018neural,feinberg2018model} and model-free methods~\cite{williams1992simple, sutton1999policy, mnih2016asynchronous,schulman2015trust,schulman2017proximal,lillicrap2015continuous, fujimoto2018addressing}. 
Among model-free methods, the evolution from Vanilla Policy Gradient~\cite{williams1992simple, sutton1999policy} to 
TRPO~\cite{schulman2015trust} 
and PPO~\cite{schulman2017proximal} has influenced the development of GRPO. 
In the context of fine-tuning LLMs, another line of work is RL from human feedback (RLHF). An early example of connecting RL with LLMs dates back to OpenAI’s work on integrating human preferences to improve text generation tasks, such as summarization using the PPO algorithm~\cite{ziegler2019fine}. This approach was later extended to fine-tune LLMs for instruction following and/or alignment on helpfulness and harmlessness~\cite{ouyang2022training, bai2022training, grattafiori2024llama, yang2024qwen2}. Due to the high data requirements and training costs of standard RLHF, off-policy methods like DPO~\cite{rafailov2023direct} and its variants~\cite{azar2024general,ethayarajh2024kto,xu2024contrastive,meng2024simpo,guo2025discriminative}, have been proposed to reduce reliance on explicit reward models. Another line of on-policy algorithms for RLHF, such as RLOO~\cite{ahmadian2024back}, ReMax~\cite{li2023remax}, and REINFORCE++~\cite{hu2025reinforce}, has been introduced to reduce the computational burden by removing the critic network in PPO. While some works~\cite{luong2024reft, kazemnejad2024vineppo, hu2025reinforce,chen2024step, xie2024monte} attempt to adapt RLHF techniques for reasoning tasks, they have not yielded significant improvements.

\textbf{Discriminative Learning.} Parallel to RL, discriminative learning is another classical learning paradigm, that has been studied extensively for many traditional tasks, including multi-class classification~\cite{cortes1995support,10.5555/944790.944813,bishop2006pattern}, AUC maximization~\cite{yang2022aucmaximizationerabig,yuan2021large}, and learning to rank~\cite{listnet,freund2003efficient,burges2005learning}. These methods are grounded in the common principle of increasing prediction scores for positive (relevant) labels (data) while decreasing scores for negative (irrelevant) ones. Nevertheless, discriminative learning remains under-explored in the cotext of LLM training. Recently, Guo et al.~\cite{guo2025discriminativefinetuninggenerativelarge} proposed discriminative probabilistic approaches for supervised fine-tuning of LLMs. However, unlike our approach, they did not employ an RL framework with verifiable rewards to fine-tune LRMs.

\section{Preliminaries}

We consider fine-tuning a generative reasoning model $\pi_\theta$ parameterized by $\theta$. The old model in one step of learning is denoted by $\po$. It is used to generate answers for a set of input questions. Given a question $q$ (with prompt included), the generated output $o$ follows the distribution $\po(\cdot|q)$, which includes reasoning traces and the final answer. Specifically, output $o$ is generated token by token, i.e., $o_t \sim \po(\cdot|q,o_{<t})$, for $t=1, \cdots, |o|$. 
We consider a rule-based reward mechanism that returns a binary value for a given question $q$ and its corresponding answer in the output $o$, which uses either exact match against extracted answer or a formal verification tool~\cite{guo2025deepseek,lambert2025tulu3pushingfrontiers,ren2025deepseekproverv2advancingformalmathematical}. Let $r(o|q)\in\{1,0\}$ denote the reward assigned to an output $o$ with respect to the input $q$. Let $p(q) = \E_{o\sim \po(\cdot|q)}[r(o|q)]\in [0,1]$, which quantifies the difficulty of the question $q$ under the model $\po$. We denote by $\po^+(\cdot|q)$ the conditional distribution of outputs when the reward is one (i.e., positive answers) and by $\po^-(\cdot|q)$ the conditional distribution of outputs when the reward is zero (i.e., negative answers). By the law of total expectation, for any function $g(o,q)$ we have
\begin{align}\label{eqn:dec}
&\E_{o\sim \po(\cdot|q)}[g(o, q)]=p(q)\E_{o\sim \po^+(\cdot|q)}[g(o,q)] +(1-p(q))\E_{o\sim \po^-(\cdot|q)}[g(o,q)].
\end{align}

\noindent{\bf Group Relative Policy Optimization (GRPO).} 
The key idea of GRPO is to generate multiple outputs for an input $q$ and define a group relative advantage function.  For  analysis, we consider the expectation formulation instead of empirical average of the GRPO  objective for maximization:
\begin{align}\label{eqn:grpo}
     &\mathcal{J}_{\text{GRPO}}(\theta)= \E_q\E_{o\sim\po(\cdot|q)}\bigg[\frac{1}{|o|}\sum_{t=1}^{|o|}f\left(\frac{\pi_\theta(o_t|q,o_{<t})}{\po(o_t|q,o_{<t})}, A(o|q)\right)\bigg] - \beta \D_\text{KL}(\pi_{\theta}||\pi_{\text{ref}}),
\end{align}
where  $f(x,y) = \min (xy, \text{clip}(x, 1-\epsilon, 1+\epsilon)y)$, 
    $A(o|q) = \frac{(r(o|q)-\E_{o'\sim\po(\cdot|q)}r(o'|q))}{\sqrt{\text{Var}_{o'\sim\po(\cdot|q)}r(o'|q)}}$ is the advantage function that quantifies how much better the reward of $o$ is compared to average reward,   $\pi_{\text{ref}}$ is a frozen reference model. 

Recently, several variants of GRPO have been introduced~\cite{yu2025dapo, chu2025gpg, lin2025cppo,zhang2025grpo,liu2025understanding}. Many of them retain the advantage function $A(o|q)$ while modifying other components such as hyper-parameter $\epsilon$, the normalization factor and the likelihood ratio. Several works employ an unnormalized advantage function $\hat A(o|q) = r(o|q)-\E_{o'\sim\po(\cdot|q)}r(o'|q)$~\cite{liu2025understanding, chu2025gpg}.

\section{Analysis of GRPO and its variants}
In the following analysis we assume $p(q)\in(0,1)$; otherwise we can remove them from consideration as done in practice~\cite{guo2025deepseek, deepscaler2025,shao2024deepseekmath}. 
\begin{prop}\label{prop:1}
Let us consider the objective of GRPO and its variants with the following form: 
\begin{align}\label{eqn:grpo}
     &\mathcal{J}_0(\theta)= \E_q\E_{o\sim\po(\cdot|q)}\bigg[\frac{1}{|o|}\sum_{t=1}^{|o|}f\left(\frac{\pi_\theta(o_t|q,o_{<t})}{\po(o_t|q,o_{<t})}, A(o|q)\right)\bigg] .
\end{align}
Assume that $f(x, y)$ is non-decreasing function of $x$ such that $f(x, y)=\I(y>0)y f^+(x, 1) - \I(y\leq 0) |y|f^-(x, 1)$, where both $f^+, f^-$ are non-decreasing functions of $x$, then we have
\begin{align}
\mathcal{J}_0(\theta)=\E_q\sqrt{p(q)(1-p(q))}\E_{o\sim\po^+(\cdot|q), o'\sim\po^-(\cdot|q)}[s_\theta^+(o, q)-s_\theta^-(o', q)],
\end{align}
where $s_{\theta}^+(o, q)  = \frac{1}{|o|}\sum_{t=1}^{|o|}f^+\left(\frac{\pi_\theta(o_t|q,o_{<t})}{\po(o_t|q,o_{<t})}, 1\right)$ and $s_{\theta}^-(o, q) = \frac{1}{|o|}\sum_{t=1}^{|o|}f^-\left(\frac{\pi_\theta(o_t|q,o_{<t})}{\po(o_t|q,o_{<t})}, 1\right)$. In particular, for GRPO we have
\begin{align}\label{eqn:grpos}
f^+(x,1) = \min(x, 1+\epsilon), \quad f^-(x,1) =\max(x, 1-\epsilon). 
\end{align}
\end{prop}
{\bf Remark:} The assumption of  $f(x, y)$ indeed holds for GRPO and its variants.  We will present the analysis for several variants of GRPO in Appendix~\ref{sec:aog}.


The proof of the above proposition is included in Appendix~\ref{app:prop} and is inspired by~\cite{mroueh2025reinforcementlearningverifiablerewards} with differences that lead to two {\bf new insights} from Proposition~\ref{prop:1} regarding the two components of $\mathcal J_0$. First, let us consider  the component $\E_{o\sim\po^+(\cdot|q), o'\sim\po^-(\cdot|q)}[s_\theta^+(o, q)-s_\theta^-(o', q)]$. Since both $f^+$ and $f^-$ are non-decreasing functions of the first argument, then both $s_\theta^+(o, q)$ and $s_\theta^-(o, q)$ are non-decreasing functions of $\pi_{\theta}(o_t|q, o_{<t})$. Hence, maximizing $\mathcal J_0$ would increase the likelihood of tokens in the positive answers and decrease the likelihood of tokens in the negative answers.  This makes sense as we would like the new model to have a high likelihood of generating a positive (correct) answer and a low likelihood of generating a negative (incorrect) answer.  This mechanism is closely related to traditional discriminative methods of supervised learning in the context of  AUC maximization~\cite{yang2022auc}, which aims to maximize the scores of positive samples $o\sim \po^+(\cdot|q)$ while minimizing scores of negative samples $o'\sim \po^-(\cdot|q)$,  where the $q$ acts like the classification task in the AUC maximization.  Hence, in the context of discriminative learning, we refer to $s^+(o, q)$ and $s^-(o,q)$ as scoring functions. Therefore, $\E_{o\sim\po^+(\cdot|q), o'\sim\po^-(\cdot|q), }[s^+(o, q)-s^-(o', q)]$ is a discriminative objective. 

\begin{figure}[!t]
  \centering
  \subfigure[]
  {\includegraphics[width=.24\textwidth]{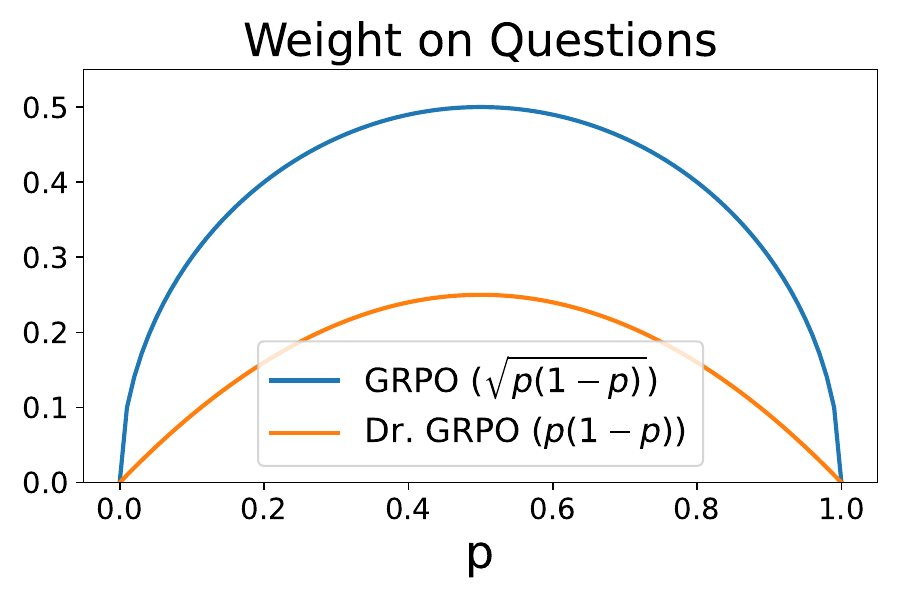}}
  \subfigure[]
  {\includegraphics[width=.24\textwidth]{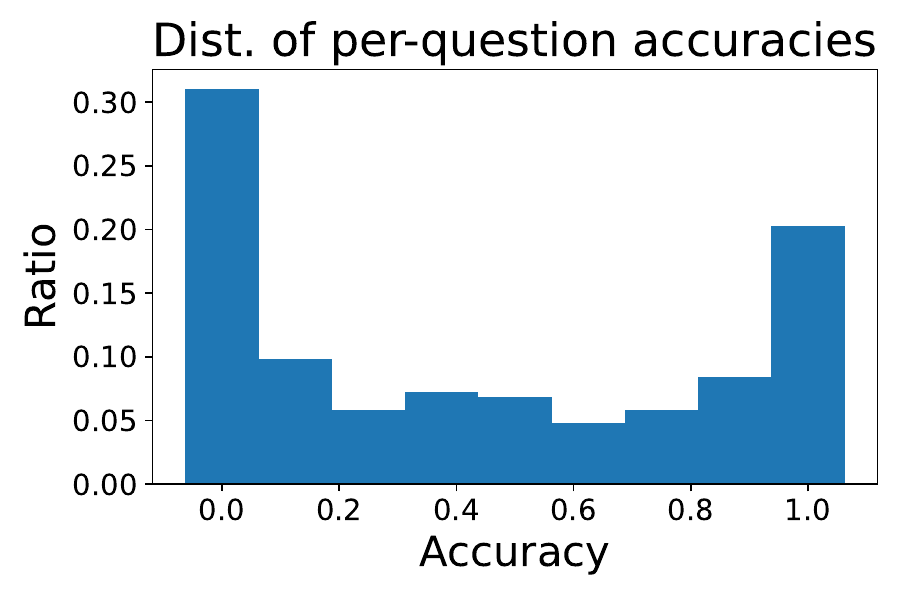}}
  \subfigure[]
  {\includegraphics[width=.24\textwidth]{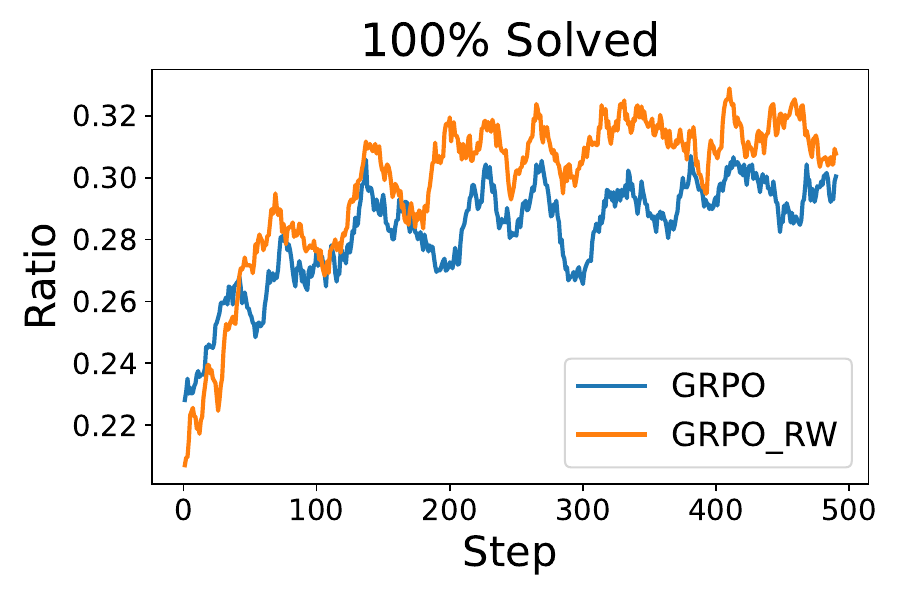}}
  \subfigure[]
  {\includegraphics[width=.24\textwidth]{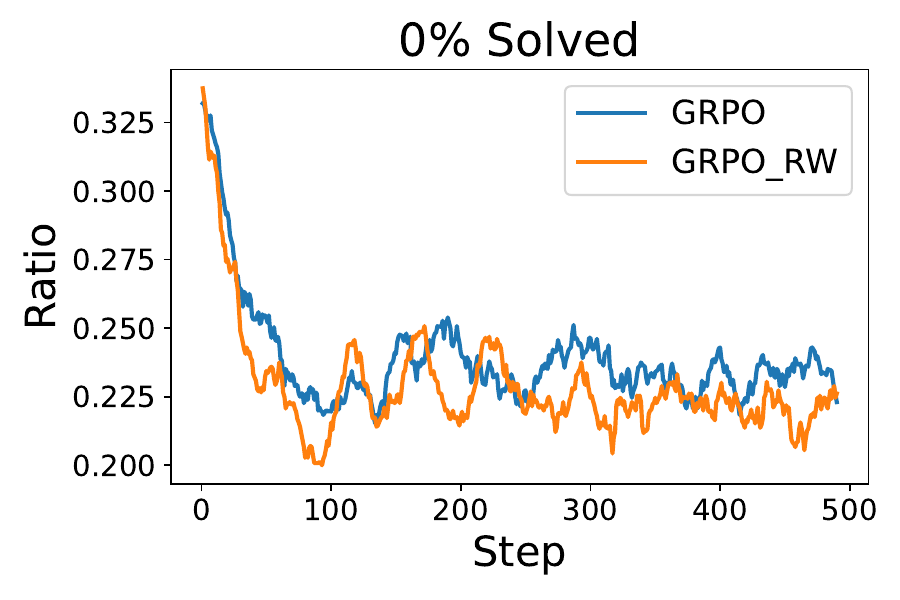}}
    \vspace*{-0.1in}
  \caption{(a) Weight on questions based on correctness probability $p$; (b) Histogram of per-question accuracy evaluated in the GRPO learning; (c) Comparison of the ratio of questions with 100\% correctness probability; (d) Comparison of the ratio of questions with 0\% correctness probability. }
  \label{fig:intro}
  \vspace*{-0.1in}
\end{figure}

Second, let us consider the component $\omega(q)=\sqrt{p(q)(1-p(q))}$, which acts like a weight scaling the discriminative objective for each individual input question. It is this component that leads to difficulty bias.    As shown in Figure~\ref{fig:intro}(a), questions with very high $p(q)$ values (close to 1) or very low $p(q)$ values (close to 0) receive small weights for their discriminative objectives, causing the optimization to focus primarily on questions of intermediate difficulty while paying little attention to hard questions ($p(q) \approx 0$) and easy questions ($p(q) \approx 1$). This mechanism may significantly hinder the learning efficiency. Intuitively, if the generated answers have only one correct solution out of 10 trials, i.e. $p(q)=0.1$, we should grasp this chance to enhance the model instead of overlooking it.  On the other hand, even when we encounter an easy question with a probability of $p(q)=0.9$, we should keep improving the model rather than being satisfied because it still makes mistakes with respect to this question. Our hypothesis is that removing this weight could accelerate the training. To validate this hypothesis, we conducted a series of empirical experiments for fine-tuning a 1.5B model as described in Section~\ref{sec:exp}. We start by examining whether a substantial number of questions have correctness probabilities ($p(q)$) near 0 or 1.  As shown in Figure~\ref{fig:intro}(b), during GRPO training, the correctness probabilities across individual questions appear broadly distributed, with many near 0 or 1. Then, we compare the original GRPO  with a variant that removes weight $\sqrt{p(q)(1-p(q))}$:
\begin{equation}\label{eqn:grpo_rw}
\begin{aligned}
    &\mathcal{J}_{\text{GRPO}\_\text{RW}} = \E_q\E_{o\sim\po^+,o'\sim\po^-}[s^+(o,q) - s^-(o',q)]- \beta \D_\text{KL}(\pi_{\theta}||\pi_{ref}).
\end{aligned}
\end{equation}
The results are shown in Figure~\ref{fig:intro}(c) and ~\ref{fig:intro}(d). We can observe that the variant without the weighting mechanism quickly achieves a higher ratio of 100\% correctness and a lower ratio of 0\% correctness, confirming the detrimental impact of the inappropriate weighting. 

We note that the difficulty bias has been pointed out in a recent work Dr. GRPO~\cite{liu2025understanding}.  To mitigate this issue, Dr. GRPO uses the un-normalized advantage function $\hat A(o|q)$. However, with a similar analysis as above (cf. Appendix~\ref{sec:aog}), we can derive that Dr. GRPO still has a question-level weight  $\omega(q) = p(q)(1-p(q))$ before the discriminative objective. As shown in Figure~\ref{fig:intro}(a), this weight mitigates but does not eliminate the imbalanced weight across questions.


\section{A Discriminative Constrained Optimization Framework}
While the last section has suggested a tangible remedy to address the difficulty bias of GRPO and its variants by removing the weight before the discriminative objective, there are other issues of the scoring function of GRPO and its variants. Next, we propose a general discriminative learning framework for reinforcing LRMs and incorporate advanced techniques to facilitate the learning.

\subsection{A basic approach} 
Motivated by the connection with AUC maximization, we redesign the objective directly from the principle of discriminative learning. For a given question $q$, let $s_\theta(o, q)$ denote a scoring function that measures how likely the model $\pi_\theta$ ``predicts'' the output $o$ for a given input $q$~\footnote{in the context of generative models, ``predicts'' is like ``generates''. }. Then the AUC score for the ``task'' $q$ is equivalent to $\E_{o\sim\po^+,o'\sim\po^-}[\I(s_\theta(o,q)> s_\theta(o',q))]$. Using a continuous surrogate function $\ell$, we form the following objective (in expectation form) for maximization:
\begin{equation}\label{eqn:diso_c}
\begin{aligned}
   \mathcal J_{1}(\theta) = \E_q\E_{o\sim\po^+(\cdot|q),o'\sim\po^-(\cdot|q)} \ell(s_\theta(o, q) - s_\theta(o', q)).
\end{aligned}
\end{equation}
Different surrogate functions $\ell(\cdot)$ can be used. For comparison, with GRPO, we simply use the identity function $\ell(s) = s$. One difference from the discriminative objective~(\ref{eqn:grpo_rw}) is that we use a single scoring function $s_\theta(o, q)$ for both positive outputs $o$ and negative outputs $o'$. It is notable that the different scoring functions for positive and negative outputs in~(\ref{eqn:grpo_rw}) actually arise from the clipping operations of GRPO objective. Recent works have found that the clipping could lead to entropy collapse~\cite{yu2025dapo}. In addition, the clipping could cause the vanishing gradient, which may also slow down the learning process. To avoid these issues, we consider non-clipping scoring functions. 

{\bf Scoring functions.} We consider two choices of scoring functions, i.e., log-likelihood and likelihood ratio. 
The log-likelihood (log-L) scoring function is defined by
$s_\theta(o, q) = \frac{1}{|o|}\sum_{t=1}^{|o|}\log \pi_{\theta}(o_t|q, o_{<t})$.
The likelihood ratio (L-ratio) scoring function is computed by
$s_\theta(o, q) = \frac{1}{|o|}\sum_{t=1}^{|o|}\frac{\pi_{\theta}(o_t|q, o_{<t})}{\po(o_t|q, o_{<t})}$. 
In Appendix~\ref{sec:connection}, we discuss the connection between the two scoring functions and the surrogate objectives of vanilla policy gradient methods~\cite{williams1992simple} and  TRPO~\cite{schulman2015trust}, respectively. 


{\bf Stabilize the training with Constrained Optimization.}  Training instability is a long-standing issue in  RL~\cite{schulman2015trust,schulman2017proximal}. Different methods have been introduced to ensure stability. Recent RL-based methods for learning reasoning models either follow the clipping operation of PPO~\cite{schulman2017proximal} or use the KL divergence regularization  $\D_\text{KL}(\pi_{\theta}||\pi_{\text{ref}})$ or $\D_\text{KL}(\po||\pi_{\theta})$~\cite{shao2024deepseekmath, su2025trust, ouyang2022training}. However, the clipping operation could lead to the entropy collapse~\cite{yu2025dapo}, which we try to avoid by using the non-clipped scoring function. The KL divergence regularization $\D_\text{KL}(\pi_{\theta}||\pi_{\text{ref}})$ while being used in traditional RL is not effective for preventing entropy collapse~\cite{skywork-or1-2025,yu2025dapo}. Regarding the regularization with $\D_\text{KL}(\po||\pi_{\theta})$, earlier studies \cite{schulman2015trust, schulman2017proximal} has found that it would be difficult to choose a single value of the regularization parameter that performs well across different problems or even within a single problem where the the characteristics change over the course of learning. To tackle this issue, we revisit the idea of trust region constraint of TRPO~\cite{schulman2015trust}, i.e., restricting the updated model $\theta$ in the trust region $\D_\text{KL}(\po||\pi_{\theta})\leq \delta$.  As a result, we solve the following {\bf discriminative constrained optimization} problem: 
\begin{equation}\label{eqn:diso}
\begin{aligned}
   &\max_{\theta} \mathcal J_{1}(\theta): = \E_q\E_{o\sim\po^+(\cdot|q),o'\sim\po^-(\cdot|q)} \ell(s_\theta(o, q) - s_\theta(o', q))\\
    & s.t. \quad \D_\text{KL}(\po||\pi_{\theta}) \leq \delta.
\end{aligned}
\end{equation}
For sake of efficiency, we use a different optimization approach from TRPO to solve the above constrained optimization. Inspired by the recent advances of non-convex ineqaulity constrained optimization algorithm~\cite{li2024model}, we adopt a squared-hinge penalty function for the constraint and solve the following problem with an appropriate penalty parameter $\beta$:
\begin{equation}\label{eqn:diso-p}
\begin{aligned}
   &\max_{\theta}  \E_q\E_{o\sim\po^+(\cdot|q),o'\sim\po^-(\cdot|q)} \ell(s_\theta(o, q) - s_\theta(o', q)) - \beta [\D_\text{KL}(\po||\pi_{\theta})- \delta]_+^2,
\end{aligned}
\end{equation}
where $[\cdot]_+ = \max\{\cdot, 0\}$.  It has been shown that under an appropriate assumption regarding the constraint function and $\beta$, solving the above squared-hinge penalized objective~(\ref{eqn:diso-p}) can return a KKT solution of the original constrained problem~(\ref{eqn:diso}). We refer the readers to~\cite{li2024model} for more in-depth analysis of this approach. 

Finally, we would like to emphasize the difference between using the squared-hinge penalty function and the regular KL divergence regularization $\beta\D_\text{KL}(\po||\pi_{\theta})$.  The squared-hinge penalty function has a dynamic weighting impact for the gradient, $ \nabla\beta [\D_\text{KL}(\po||\pi_{\theta})- \delta]_+^2 =  2\beta [\D_\text{KL}(\po||\pi_{\theta})- \delta]_+\nabla \D_\text{KL}(\po||\pi_{\theta})$, such that if the constraint is satisfied then the weight $2\beta [\D_\text{KL}(\po||\pi_{\theta})- \delta]_+$ before the gradient of the regularization term $ \D_\text{KL}(\po||\pi_{\theta})$ becomes zero. This means the KL divergence regularization is only effective when the constraint is violated.  In contrast, the regular KL divergence regularization $\beta\D_\text{KL}(\po||\pi_{\theta})$ always contributes a gradient  $\beta\nabla\D_\text{KL}(\po||\pi_{\theta})$ no matter whether the constraint is satisfied or not, which could harm the learning. 


\subsection{An improved approach for tackling imbalanced rollouts}
One advantage of designing the objective based on the principle of discriminative learning is the ability to leverage a wide range of advanced techniques from the literature to improve training. A key challenge in RL fine-tuning for reasoning models is the sparse rewards, which lead to imbalance in generated rollouts. Specifically, for some questions where $p(q) \ll 1$, the number of negative outputs can significantly exceed the number of positive ones. This reflects a classic data imbalance issue, which has been extensively studied in the discriminative learning community~\cite{pmlr-v162-zhu22g,DBLP:journals/corr/Shalev-ShwartzW16,10.5555/3294996.3295057}. To address this issue, we consider distributionally robust optimization (DRO)~\cite{pmlr-v162-zhu22g,10.5555/3294996.3295057}.  

Let us first discuss why the basic approach could be ineffective for combating the imbalanced rollouts. The objective function $\mathcal J_1$ is motivated by maximizing AUC for each question $q$, i.e.,  $\E_{o\sim\po^+,o'\sim\po^-}[\I(s_\theta(o,q)> s_\theta(o',q))]$. However, when there is much more negative data than positive data, AUC is not a good measure. For example, let us consider a scenario where there are 1 positive $o_+$ and 100 negatives $\{o_-^1,\ldots, o_-^{100}\}$. If the scores of these data are $s(o^{1}_-,q)=0.9, s(o_+,q)=0.5, s(o^{2}_-,q)=s(o^{3}_-,q)\ldots=s(o^{100}_-,q)=0.001$, then the AUC score is $\frac{99}{100}=0.99$. The AUC score is high but is not informative as the model still generates the negative data $o^1_-$ more likely than the positive data $o_+$. In the literature, this issue has been addressed by maximizing a partial AUC score, which considers the pairwise order between all positives and the top ranked negatives. We utilize a surrogate function of partial AUC score formulated from the perspective of DRO~\cite{pmlr-v162-zhu22g}. 

Consider a question $q$ and a positive data $o$. We denote by $\mathcal Q$ the set of probability measures $Q$ on negative data given $q$ (absolutely continuous
with respect to $\po^-(\cdot|q)$).  Denote by $\D_{\text{KL}}(Q, \po^-(\cdot|q))$ the KL divergence between a distribution $Q$ and the negative data distribution $\po^-(\cdot|q)$. A DRO formulation for partial AUC maximization is given by~\cite{pmlr-v162-zhu22g}[Theorem 2]:
\begin{align*}
\inf_{Q\in\mathcal Q}\tau\D_{\text{KL}}&(Q, \po^-(\cdot|q))  + \E_{o'\sim Q}[s_\theta(o, q) - s_{\theta}(o', q)] :\\
& = - \tau \log\bigg(\E_{o'\sim \po^-(\cdot|q)}\exp\bigg(\frac{s_{\theta}(o', q)-s_\theta(o, q)}{\tau}\bigg)\bigg). 
\end{align*}
As a result, we construct the following DRO-based objective for maximization:
\begin{align}
\mathcal J_2(\theta) = -\E_{q}\E_{o\sim \po^+(\cdot|q)} \tau \log\bigg(\E_{o'\sim \po^-(\cdot|q)}\exp\bigg(\frac{s_{\theta}(o', q)-s_\theta(o, q)}{\tau}\bigg)\bigg). 
\end{align}
It is easy to show that $\mathcal J_2(\theta)\leq \mathcal J_1(\theta)$ by Jensen's inequality for the convex function $-\log$. Hence, maximizing $\mathcal J_2(\theta)$ will automatically increasing $\mathcal J_1(\theta)$. However, the reverse is not true. This also explains why maximizing $\mathcal J_2(\theta)$ could be more effective than maximizing $\mathcal J_1(\theta)$. The above risk function is also known as optimized certainty equivalents (OCE) in mathematical finance~\cite{oce}. We would like to point out that although OCE or DRO has been considered for RL in existing works~\cite{wang2025reductionsapproachrisksensitivereinforcement,xu2025distributionallyrobustdirectpreference},  they differ from our work for addressing different issues. Wang et al.~\cite{wang2025reductionsapproachrisksensitivereinforcement} apply the OCE to compute a robust reward, replacing the standard expected reward used in traditional RL settings. Xu et al.~\cite{xu2025distributionallyrobustdirectpreference} adopt DRO for direct preference optimization of LLMs, aiming to mitigate the noise in human preference data by addressing the distributional shift between the empirical distribution of $(q, o, o')$ and its true underlying distribution.

\begin{table}[t]
    \centering
        \caption{Comparison of different methods for reinforcing large reasoning models. ``L-ratio'' means likelihood ratio, ``log-L'' means log-likelihood, ``proper'' means any proper scoring function.}
    \label{tab:comp}
    \resizebox{0.98\linewidth}{!}{
    \begin{tabular}{c|c|c|c|c|c}
    \toprule
         Method & Difficulty Bias & Clipping & KL Divergence & Score Function &Tackles Imbalanced Rollouts\\
    \midrule
         GRPO~\cite{guo2025deepseek} &  Yes & Yes & regularization, $\pi_{\text{ref}}$& clipped L-ratio& No \\
         Dr. GRPO~\cite{liu2025understanding} &  Yes & Yes &  No & clipped L-ratio &No \\
         DAPO~\cite{yu2025dapo} &  Yes & Yes &No & clipped L-ratio & No\\
         GPG~\cite{chu2025gpg} &  Yes & No &No & log-L &No\\
         TRPA~\cite{su2025trust} &  No & No & regularization, $\pi_{\text{old}}$ & log L-ratio & No\\
    \midrule
         DisCO &  No & No & constraint, $\pi_{\text{old}}$ & proper & Yes\\
    \bottomrule
    \end{tabular}
    }
\vspace*{-0.1in}
\end{table}
Finally, we solve the following discriminative constrained optimization problem by using the same squared-hinge penalty method: 
\begin{equation}\label{eqn:diso2}
\begin{aligned}
   &\max_{\theta} \mathcal J_{2}(\theta): = -\E_{q}\E_{o\sim \po^+(\cdot|q)} \tau \log\bigg(\E_{o'\sim \po^-(\cdot|q)}\exp\bigg(\frac{s_{\theta}(o', q)-s_\theta(o, q)}{\tau}\bigg)\bigg),\\
    & s.t. \quad \D_\text{KL}(\po||\pi_{\theta}) \leq \delta.
\end{aligned}
\end{equation}
To differentiate the approach for solving~(\ref{eqn:diso}) and~(\ref{eqn:diso2}), we refer to the former as {\bf DisCO-b} and the latter as {\bf DisCO}. In practice, all expectations will be replaced by empirical averages and the KL divergence is also estimated at each iteration by using sampled data following~\cite{ouyang2022training}. We present a full algorithm in Algorithm~\ref{alg:disco} in Appendix.  Finally, we give a comparison between DisCO and existing RL fine-tuning methods for reinforcing LRMs from different aspects in Table~\ref{tab:comp}.

\section{Experiments}\label{sec:exp}

In this section, we empirically evaluate the effectiveness of the proposed DisCO by comparing with GRPO and other variants for reinforcing SFT-finetuned models.

\begin{table}[t]
  \centering
  \caption{ Comparison with baseline models and baseline methods for fine-tuning 1.5B models. OpenAI-o1-preview is included as a reference.  MRL denotes Max Response Length utilized in training/testing. The shaded models are trained by other works and the shaded numbers are reported in their original works or in~\cite{deepscaler2025}. All other results are either evaluated on existing models or on the models trained by us using different approaches. Methods in the bottom area are all for fine-tuning  DeepSeek-R1-Distill-Qwen-1.5B model on the same DeepScaleR dataset. DS is short for DeepSeek-R1, DSR is short for DeepScaleR.}
  \resizebox{\linewidth}{!}{
    \begin{tabular}{lc|cccccc|c}
    \toprule
    Model/Method & MRL(Train/Test) & AIME 2024 & AIME 2025 & MATH 500 & AMC 2023 & Minerva & O-Bench & Avg. \\
    \midrule
     \cellcolor[rgb]{ .851,  .851,  .851} OpenAI-o1-Preview & -     & \cellcolor[rgb]{ .851,  .851,  .851} 0.4   & -     &  \cellcolor[rgb]{ .851,  .851,  .851}0.814 & -     & -     & -     & - \\
    \cellcolor[rgb]{ .851,  .851,  .851} DS-Distill-Qwen-1.5B & 32k+   /   32k &  \cellcolor[rgb]{ .851,  .851,  .851} 0.288 & \cellcolor[rgb]{ 1,  1,  1}0.263 &  \cellcolor[rgb]{ .851,  .851,  .851} 0.828 &  \cellcolor[rgb]{ .851,  .851,  .851} 0.629 &  \cellcolor[rgb]{ .851,  .851,  .851} 0.265 &  \cellcolor[rgb]{ .851,  .851,  .851} 0.433 & \cellcolor[rgb]{ 1,  1,  1}0.451 \\
    \cellcolor[rgb]{ .851,  .851,  .851} DS-Distill-Qwen-1.5B & 32k+   /   8k & \cellcolor[rgb]{ 1,  1,  1}0.181 & \cellcolor[rgb]{ 1,  1,  1}0.215 & \cellcolor[rgb]{ 1,  1,  1}0.758 & \cellcolor[rgb]{ 1,  1,  1}0.515 & \cellcolor[rgb]{ 1,  1,  1}0.237 & \cellcolor[rgb]{ 1,  1,  1}0.353 & \cellcolor[rgb]{ 1,  1,  1}0.376 \\
    \cellcolor[rgb]{ .851,  .851,  .851} STILL-3-1.5B-preview & 29k   /   32k & \cellcolor[rgb]{ .851,  .851,  .851}0.325 & \cellcolor[rgb]{ 1,  1,  1}0.248 & \cellcolor[rgb]{ .851,  .851,  .851}0.844 & \cellcolor[rgb]{ .851,  .851,  .851}0.667 & \cellcolor[rgb]{ .851,  .851,  .851}0.290 & \cellcolor[rgb]{ .851,  .851,  .851}0.454 & \cellcolor[rgb]{ 1,  1,  1}0.471 \\
    \midrule
    \cellcolor[rgb]{ .851,  .851,  .851} DSR-1.5B-Preview & 24k   /   32k & \cellcolor[rgb]{ .851,  .851,  .851}\textbf{0.431} & \cellcolor[rgb]{ 1,  1,  1}0.304 & \cellcolor[rgb]{ .851,  .851,  .851}\textbf{0.878} & \cellcolor[rgb]{ .851,  .851,  .851}0.736 & \cellcolor[rgb]{ .851,  .851,  .851}0.302 & \cellcolor[rgb]{ .851,  .851,  .851}0.500 & \cellcolor[rgb]{ 1,  1,  1}0.525 \\
    \cellcolor[rgb]{ .851,  .851,  .851} DSR-1.5B-Preview & 24k   /   8k & \cellcolor[rgb]{ 1,  1,  1}0.358 & \cellcolor[rgb]{ 1,  1,  1}0.258 & \cellcolor[rgb]{ 1,  1,  1}0.860 & \cellcolor[rgb]{ 1,  1,  1}0.679 & \cellcolor[rgb]{ 1,  1,  1}0.297 & \cellcolor[rgb]{ 1,  1,  1}0.473 & \cellcolor[rgb]{ 1,  1,  1}0.488 \\
    GRPO  & 8k   /   8k & 0.277 & 0.242 & 0.838 & 0.647 & 0.276 & 0.462 & 0.457 \\
    GRPO-ER & 8k   /   8k & 0.298 & 0.242 & 0.839 & 0.649 & 0.279 & 0.452 & 0.460 \\
    Dr. GRPO & 8k   /   8k & 0.252 & 0.238 & 0.831 & 0.631 & 0.268 & 0.440 & 0.443 \\
    DAPO  & 8k   /   8k & 0.310 & 0.252 & 0.848 & 0.675 & 0.296 & 0.456 & 0.473 \\
    TRPA  & 8k   /   8k & 0.354 & 0.235 & 0.835 & 0.653 & 0.283 & 0.458 & 0.470 \\
    DisCO (L-ratio) & 8k   /   8k & 0.381 & 0.306 & \textbf{0.878} & 0.746 & 0.319 & \textbf{0.512} & 0.524 \\
    DisCO (log-L) & 8k   /   8k & 0.404 & \textbf{0.317} & 0.876 & \textbf{0.758} & \textbf{0.333} & 0.509 & \textbf{0.533} \\
    \bottomrule
    \end{tabular}%
    }
  \label{tab:1.5B}%
  \centering
  \caption{ Comparison with baseline models and baseline methods for fine-tuning 7B models. Methods in the bottom area are all for fine-tuning  DeepSeek-R1-Distill-Qwen-7B model on the same DeepScaleR dataset. }
  \resizebox{\linewidth}{!}{
    \begin{tabular}{lc|cccccc|c}
    \toprule
    Model/Method & MRL(Train/Test) & AIME 2024 & AIME 2025 & MATH 500 & AMC 2023 & Minerva & O-Bench & Avg. \\
    \midrule
    \cellcolor[rgb]{ .851,  .851,  .851} DS-Distill-Qwen-7B & 32k+   /   32k &  \cellcolor[rgb]{ .851,  .851,  .851}0.560 & \cellcolor[rgb]{ 1,  1,  1}0.396 &  \cellcolor[rgb]{ .851,  .851,  .851}0.923 &  \cellcolor[rgb]{ .851,  .851,  .851}0.825 &  \cellcolor[rgb]{ .851,  .851,  .851}0.380 &  \cellcolor[rgb]{ .851,  .851,  .851}0.568 & \cellcolor[rgb]{ 1,  1,  1}0.609 \\
    \cellcolor[rgb]{ .851,  .851,  .851} DS-Distill-Qwen-7B & 32k+  /   8k & \cellcolor[rgb]{ 1,  1,  1}0.402 & \cellcolor[rgb]{ 1,  1,  1}0.292 & \cellcolor[rgb]{ 1,  1,  1}0.873 & \cellcolor[rgb]{ 1,  1,  1}0.688 & \cellcolor[rgb]{ 1,  1,  1}0.355 & \cellcolor[rgb]{ 1,  1,  1}0.471 & \cellcolor[rgb]{ 1,  1,  1}0.513 \\
        \midrule
    \cellcolor[rgb]{ .851,  .851,  .851} GRPO-LEAD-7B & 8k   /   8k & \cellcolor[rgb]{ .851,  .851,  .851}0.470 & \cellcolor[rgb]{ .851,  .851,  .851}0.345 & \cellcolor[rgb]{ 1,  1,  1}0.893 & \cellcolor[rgb]{ 1,  1,  1}0.748 & \cellcolor[rgb]{ 1,  1,  1}0.372 & \cellcolor[rgb]{ 1,  1,  1}0.500 & \cellcolor[rgb]{ 1,  1,  1}0.555 \\
    \cellcolor[rgb]{ .851,  .851,  .851} TRPA  & 8k   /   8k & 0 \cellcolor[rgb]{ .851,  .851,  .851}.570 & -     &  \cellcolor[rgb]{ .851,  .851,  .851}0.870 & \cellcolor[rgb]{ .851,  .851,  .851} 0.780 &  \cellcolor[rgb]{ .851,  .851,  .851}0.360 &  \cellcolor[rgb]{ .851,  .851,  .851}0.550 & - \\
    GRPO  & 8k   /    8k & 0.498 & 0.394 & 0.916 & 0.807 & 0.381 & 0.555 & 0.592  \\
    GRPO-ER & 8k   /   8k & 0.515 & 0.381 & 0.916 & 0.825 & 0.376 & 0.544 & 0.593 \\
    Dr. GRPO & 8k   /   8k & 0.488 & 0.346 & 0.910 & 0.792 & 0.368 & 0.546 & 0.575  \\
    DAPO  & 8k   /   8k & 0.454 & 0.335 & 0.907 & 0.799 & 0.388 & 0.535 & 0.570 \\
    TRPA  & 8k   /   8k & 0.510 & 0.367 & 0.898 & 0.779 & 0.379 & 0.534 & 0.578 \\
    DisCO (L-ratio) & 8k   /   8k & \textbf{0.583} & \textbf{0.421} & 0.923 & 0.852 & 0.399 & 0.585 & \textbf{0.627} \\
    DisCO (log-L) & 8k   /   8k & 0.558 & 0.410 & \textbf{0.927} & \textbf{0.854} & \textbf{0.410} & \textbf{0.592} & 0.625 \\
    \bottomrule
    \end{tabular}%
    }
  \label{tab:7B}%
  \centering
  \caption{ Comparison with baseline models and baseline methods for fine-tuning 8B models. Methods in the bottom area are all for fine-tuning  DeepSeek-R1-Distill-Llama-8B model on the same DeepScaleR dataset. }
  \resizebox{\linewidth}{!}{
    \begin{tabular}{lc|cccccc|c}
    \toprule
    Model/Method & MRL(Train/Test) & AIME 2024 & AIME 2025 & MATH 500 & AMC 2023 & Minerva & O-Bench & Avg. \\
    \midrule
    \cellcolor[rgb]{ .851,  .851,  .851} DS-Distill-Llama-8B & 32k+   /   32k &  0.506 & 0.346 & 0.896 & 0.815 & 0.295 & 0.541 & 0.566 \\
    \cellcolor[rgb]{ .851,  .851,  .851} DS-Distill-Llama-8B & 32k+  /   8k & 0.348 & 0.238 & 0.825 & 0.652 & 0.267 & 0.440 & 0.462 \\
        \midrule
    GRPO  & 8k   /    8k & 0.410 & 0.240 & 0.873 & 0.759 & 0.307 & 0.506 & 0.516 \\
    GRPO+ER & 8k   /   8k & 0.408 & 0.277 & 0.882 & 0.785 & 0.311 & 0.511 & 0.529 \\
    Dr. GRPO & 8k   /   8k & 0.423 & 0.285 & 0.867 & 0.786 & 0.300 & 0.497 & 0.526 \\
    DAPO  & 8k   /   8k & 0.333 & 0.308 & 0.879 & 0.794 & 0.325 & 0.522 & 0.527 \\
    TRPA  & 8k   /   8k & 0.454 & 0.279 & 0.864 & 0.756 & 0.289 & 0.518 & 0.527 \\
    DisCO (L-ratio) & 8k   /   8k & 0.506 & \textbf{0.356} & \textbf{0.900} & 0.831 & 0.326 & 0.553 & 0.579 \\
    DisCO (log-L) & 8k   /   8k & \textbf{0.523} & 0.354 & 0.896 & \textbf{0.843} & \textbf{0.331} & \textbf{0.560} & \textbf{0.584} \\
    \bottomrule
    \end{tabular}%
    }
  \label{tab:8B}%
  \vspace*{-0.1in}
\end{table}%

\textbf{Task Setting.} We validate our method on mathematical reasoning tasks. Specifically,  we use the DeepScaleR-Preview-Dataset~\cite{deepscaler2025} for training, which includes AIME problems from 1984 to 2023, AMC problems before 2023, and questions from the Omni-MATH~\cite{gao2024omni} and Still~\cite{min2024imitate} datasets, totaling approximately 40.3k unique problem-answer pairs. We evaluate models on six benchmark datasets: AIME 2024, AIME 2025, MATH 500~\cite{hendrycks2021measuring,lightman2023let}, AMC 2023, Minerva~\cite{lewkowycz2022solving}, and Olympiad Bench (O-Bench)~\cite{he2024olympiadbench}. Following~\cite{guo2025deepseek, deepscaler2025}, we adopt the pass@1 metric~\cite{chen2021evaluating} averaged over $k=16$ responses for each question to ensure the reliability of model performances. The metric for each question is calculated as $\frac{1}{k}\sum_{i=1}^k \I(o_i \ \text{is correct})$, where $o_i$ denotes the $i$-th generated response. For both the training and evaluation of our method and the baselines (unless otherwise specified), the maximum response length is limited to 8k tokens. To verify the generalizability of our method to other datasets, we also conducted experiments on DAPO-Math-17k~\cite{yu2025dapo} dataset, which is included in the Appendix~\ref{sec:dapo17k}.

\textbf{Models.} We conduct experiments with fine-tuning three models: DeepSeek-R1-Distill-Qwen-1.5B model (Q1.5B), DeepSeek-R1-Distill-Qwen-7B model (Q7B),  and DeepSeek-R1-Distill-Llama-8B (L8B). All are distilled reasoning models.

\textbf{Baselines.} We primarily compare our methods with five most recent state-of-the-art reinforcement learning methods, including (1) GRPO~\cite{guo2025deepseek}; (2) GRPO with an entropy regularization (GRPO-ER) that adds an entropy on probabilities of output tokens as a regularization to prevent entropy collapse, which is used by DeepScaleR~\cite{deepscaler2025}; (3) Dr. GRPO~\cite{liu2025understanding}; 
(4) DAPO's objective~\cite{yu2025dapo}; 
(5) TRPA~\cite{su2025trust}.  
For a comprehensive evaluation, we also include a set of reasoning models that are trained from the same base model by other studies with various techniques, such as (6) STILL-3-1.5B-preview~\cite{chen2025empirical}, which adapt GRPO by periodically replacing the reference model after a fixed number
of training steps; (7) DeepScaleR(DSR)-1.5B-Preview that uses maximum response length of 24k for training~\cite{deepscaler2025}; 
(8) GRPO-LEAD-7B~\cite{zhang2025grpo}, which extends GRPO by incorporating length-dependent rewards, explicit penalty terms, and difficulty-based advantage reweighting to encourage concise and precise reasoning.


\textbf{Training Details.} For all the methods, we tune the constant learning rate in $[5\mathrm{e}^{-7}, 1\mathrm{e}^{-6}, 2\mathrm{e}^{-6}]$ with AdamW optimizer with weight decay as 0.01. Generally, a learning rate of $2\text{e}^{-6}$ works better for the Q1.5B model, $1\text{e}^{-6}$ for the Q7B model, and $5\text{e}^{-7}$ for the L8B model. We employ a training batch size of 128, a mini-batch size of 32, and 8 responses for each question. The temperature is set to 0.6 for both training and evaluation, following the usage recommendation from \cite{guo2025deepseek}.  For GRPO, $\beta$ is set to 0.001 as commonly used~\cite{chen2025empirical, deepscaler2025}. For GRPO-ER, we use a coefficient of 0.001 for the entropy regularization~\cite{deepscaler2025}. For DAPO, we set $\epsilon_{low}$ to 0.2 and $\epsilon_{high}$ to 0.28 by following their paper. For our method, $\delta$ is set to $10^{-4}$ based on the empirical observation that the average KL divergence is around $2*10^{-5}$ and $\beta$ is set to $10^3$ such that the effective weight of the KL regularization when the constraint is violated by $\delta$ is on the order of $\beta*\delta =0.1$. Since L-ratio and log-L scoring functions have different orders, we choose $\tau=1$ for L-ratio and $\tau=10$ for log-L scoring function, from $\{0.5, 1, 5, 10\}$. For fair comparisons, we do not implement Dynamic Sampling~\cite{yu2025dapo} for DAPO and other methods, as it introduces approximately three times the sampling cost at each training step. All methods are run for 1,400 steps on Q1.5B models and 1,000 steps on Q7B/L8B models. Evaluations are conducted every 200 steps, and the best performance for each method is reported.

\subsection{Comparison with Baselines}
\textbf{Performance.} We evaluate all the models across six mathematics-focused benchmark datasets to demonstrate the effectiveness of DisCO. The results are summarized in Table~\ref{tab:1.5B}, ~\ref{tab:7B} and~\ref{tab:8B}. From Table~\ref{tab:1.5B} for Q1.5B models, we can observe that our proposed DisCO methods consistently outperform other baselines by a large margin. Notably, DisCO (log-L) with 8k length for both training and inference achieves an 7\% average improvement over GRPO and surpasses DeepScaleR-1.5B-Preview that was trained with maximum 24k length and evaluated with 32k length. A similar trend is observed for Q7B models and L8B models (Table~\ref{tab:7B}
 and Table~\ref{tab:8B}), where DisCO significantly outperforms all competing approaches. 

\begin{figure}[!t]
  \centering
  {\includegraphics[width=.99\textwidth]{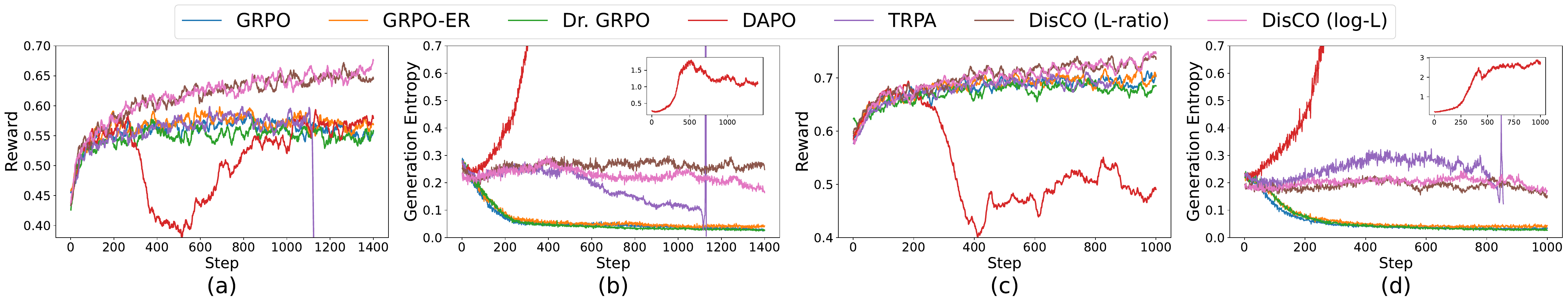}}
    \vspace*{-0.1in}
  \caption{ Training dynamics of different methods: left two are for fine-tuning DeepSeek-R1-Distill-Qwen-1.5B model and right two are for fine-tuning DeepSeek-R1-Distill-Qwen-7B model. (a), (c) plot the training reward (averaged over generated outputs for questions used in each step) vs the number of training steps (cf. Algorithm~\ref{alg:disco}); (b), (d) plot the generation entropy vs training steps. }
  \label{fig:learning}
  \vspace*{-0.1in}
\end{figure}

\textbf{Training Dynamics.} 
We compare the training dynamics of different methods in terms of training rewards and generation entropy. From Figure~\ref{fig:learning} for fine-tuning Qwen-1.5B and Qwen-7B models, we can see that all baselines suffer from premature saturation due to either entropy collapse for GRPO, GRPO-ER, Dr. GRPO or excessive entropy growth of DAPO, which leads to an early deterministic or highly random policy. The entropy collapse phenomenon is also observed by \cite{yu2025dapo, skywork-or1-2025,deepscaler2025}. TRPA that uses a KL divergence regularization is also observed with instability in the generation entropy in later steps (around 1100 for Q1.5B model and around 800 for Q7B model).  
In contrast, our methods with the two scoring functions are most stable, with training rewards kept increasing and generation entropy maintained around 0.22.  
We also include training dynamics for the L8B model in Appendix~\ref{sec:8b_train}, which follows a similar trend.

\begin{figure}[!t]
  \centering
  \subfigure
  {\includegraphics[width=.32\textwidth]{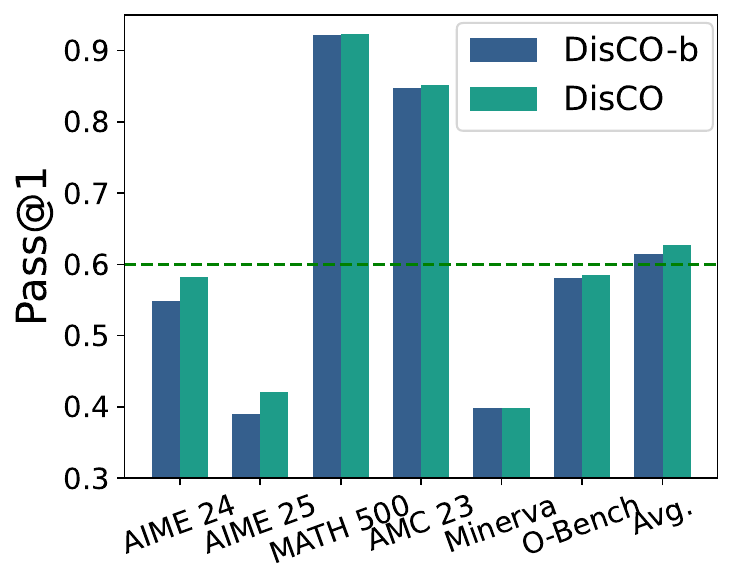}}
  \subfigure
  {\includegraphics[width=.32\textwidth]{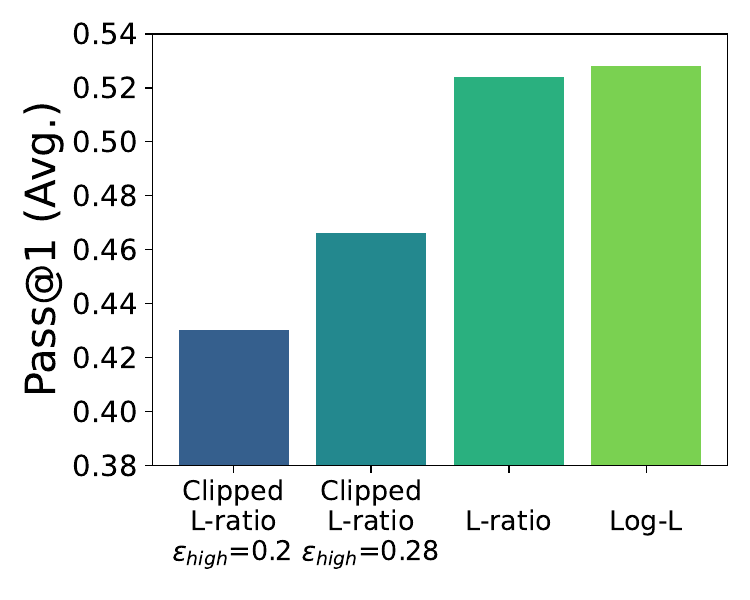}}
  \subfigure
  {\includegraphics[width=.32\textwidth]{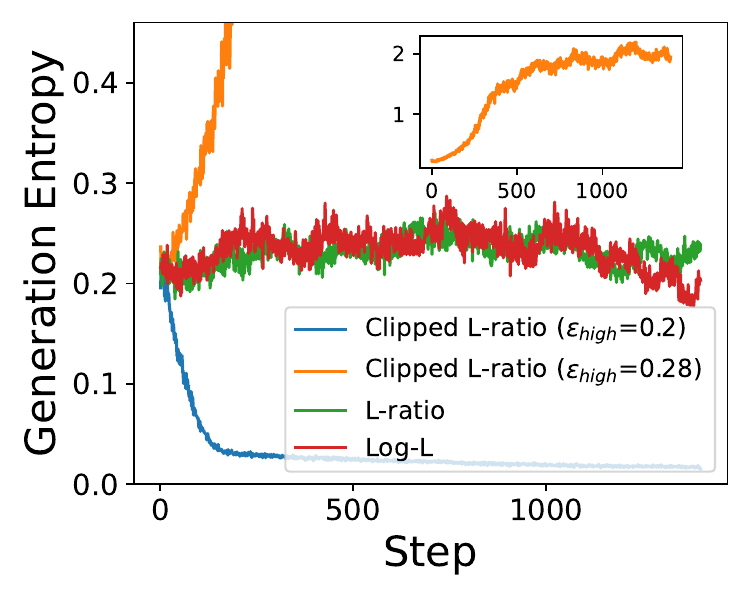}}
    \vspace*{-0.1in}
  \caption{ Ablation studies: left for comparing DisCO vs DisCO-b; middle and right for comparing clipping with non-clipping scoring functions.}
  \label{fig:score}
  \vspace*{-0.25in}
\end{figure}


\subsection{Ablation Studies}

\textbf{DisCO vs DisCO-b.} Figure~\ref{fig:score} (left) compares DisCO with DisCO-b using the L-ratio scoring function for training Q7B models for 1000 steps. The comparison clearly demonstrates the significant improvements of DisCO over DisCO-b, especially on the difficult AIME datasets. We also compare  DisCO with DisCO-b for other settings in Appendix~\ref{sec:disco-b}, and observe that DisCO is consistently better than DisCO-b on average in all settings.   It is also notable that DisCO-b with different scoring functions are also better than other baselines trained or evaluated by us in Table~\ref{tab:1.5B} and Table \ref{tab:7B}. 

\textbf{Clipping vs Non-Clipping scoring functions.} We compare non-clipping scoring functions L-ratio, log-L with clipped L-ratio~\eqref{eqn:grpos} in our DisCO-b approach for training Q1.5B models in Figure~\ref{fig:score} (middle and right). For the clipped L-ratio, we adopt two versions: one with $\epsilon_{high} = 0.2$ to align with GRPO objective, and another with $\epsilon_{high} = 0.28$ similar to DAPO objective.  We can see that clipped L-ratio with $\epsilon_{high} = 0.2$ causes the entropy collapse while clipped L-ratio with $\epsilon_{high} = 0.28$ leads to excessively high entropy level, both yielding worse performance than non-clipping scoring functions. 

\begin{figure}[!t]
  \centering
  \subfigure
  {\includegraphics[width=.32\textwidth]{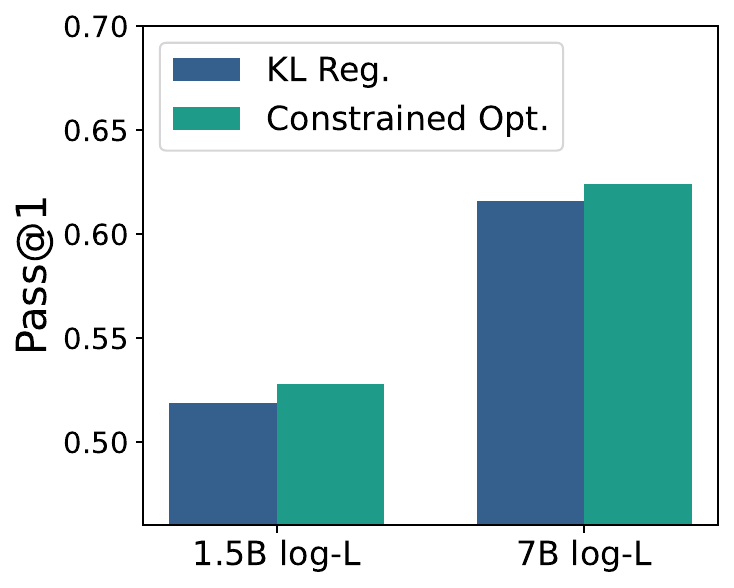}}
  \subfigure
  {\includegraphics[width=.32\textwidth]{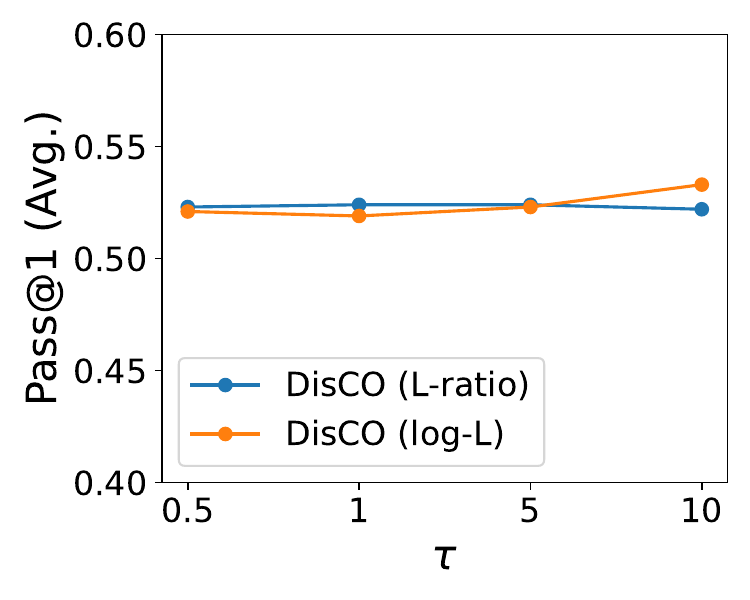}}
 \subfigure
  {\includegraphics[width=.32\textwidth]{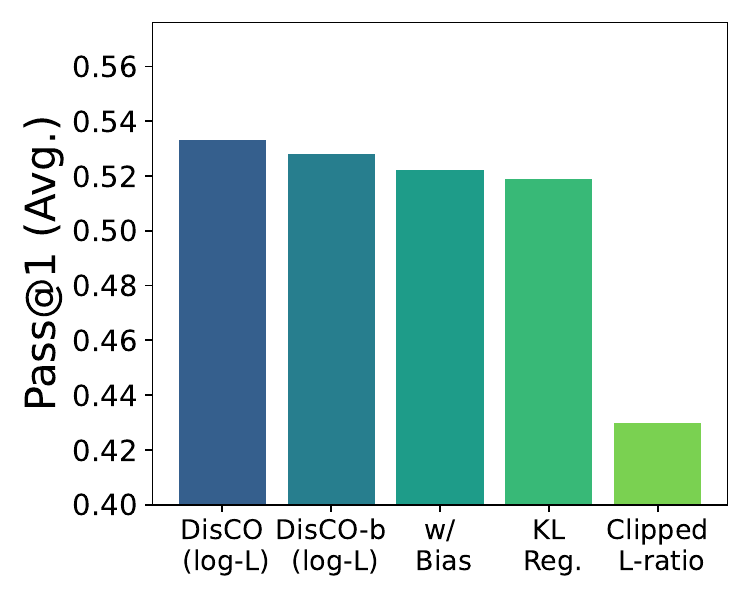}}
    \vspace*{-0.1in}
  \caption{ Ablation studies: left for comparing KL regularization vs constrained optimization; middle for sensitivity of DisCO w.r.t. the hyperparameter $\tau$; right for contribution of each component.}
  \label{fig:kl_vs_cons}
  \vspace*{-0.2in}
\end{figure}

\textbf{KL Regularization vs Constrained Optimization.} We investigate the advantages of constrained optimization over KL regularization for DisCO. Specifically,  the KL regularization weight is set to the commonly used 0.001~\cite{chen2025empirical, deepscaler2025}. As shown in Figure~\ref{fig:kl_vs_cons} (left), constrained optimization performs better than KL regularization on both Q1.5B and Q7B models. Moreover, during our experiments, we observed that KL regularization leads to instability in training on Q7B models, similar to TRPA, which indicates that KL regularization is not sufficient to stabilize training.

\textbf{Sensitivity of hyperparameter $\tau$.} We study the sensitivity of DisCO to hyperparameter $\tau$ on training Q1.5B models. Similar to above experiments, we run DisCO for 1400 steps with different $\tau\in\{0.5, 1, 5, 10\}$.  The result shown in Figure~\ref{fig:score} (right) indicates that DisCO is not sensitive to $\tau$ in these ranges. 

\textbf{Effect of each design choice.} We analyze the individual contribution of each component in DisCO by replacing its components separately with other designs. We experiment on Q1.5B models and compare with (1) DisCO-b that removes hard negative weighting; (2) adding question-level weight bias $\sqrt{p(q)(1-p(q)}$ to DisCO-b, (3) replacing the KL-divergence constraint with a KL-divergence regularization in DisCO-b, and (4) using a clipping scoring function with $\epsilon_{high} = 0.2$ in DisCO-b, respectively.   From Figure~\ref{fig:kl_vs_cons} (right), we can see that each of our proposed components is important in DisCO's improvement, where the use of a non-clipping scoring function is of vital importance. 

\section{Conclusion}
In this work, we have proposed a novel discriminative constrained optimization framework for reinforcing large reasoning models, motivated by the analysis of the GRPO objective. The proposed framework is grounded in the principle of discriminative learning, avoiding difficulty bias and enhancing training stability with constrained trust region optimization. The experiments on mathematical reasoning demonstrated the significant superiority of our approaches, compared with GRPO and its recent variants. While this work focuses on binary rewards, future extensions could incorporate discriminative ranking objectives, like~\cite{listnet}, to handle non-binary rewards. 
It would be interesting to apply the proposed approaches for fine-tuning larger models or other reasoning tasks.

\section*{Acknowledgements}
We are grateful for the reviewers' constructive comments. G. Li and T. Yang were partially supported by NSF grant \#2306572. 

\bibliography{ref}

@misc{yang2022aucmaximizationerabig,
      title={AUC Maximization in the Era of Big Data and AI: A Survey}, 
      author={Tianbao Yang and Yiming Ying},
      year={2022},
      eprint={2203.15046},
      archivePrefix={arXiv},
      primaryClass={cs.LG},
      url={https://arxiv.org/abs/2203.15046}, 
}

@InProceedings{pmlr-v162-zhu22g,
  title = 	 {When {AUC} meets {DRO}: Optimizing Partial {AUC} for Deep Learning with Non-Convex Convergence Guarantee},
  author =       {Zhu, Dixian and Li, Gang and Wang, Bokun and Wu, Xiaodong and Yang, Tianbao},
  booktitle = 	 {Proceedings of the 39th International Conference on Machine Learning},
  pages = 	 {27548--27573},
  year = 	 {2022},
  editor = 	 {Chaudhuri, Kamalika and Jegelka, Stefanie and Song, Le and Szepesvari, Csaba and Niu, Gang and Sabato, Sivan},
  volume = 	 {162},
  series = 	 {Proceedings of Machine Learning Research},
  month = 	 {17--23 Jul},
  publisher =    {PMLR},
  url = 	 {https://proceedings.mlr.press/v162/zhu22g.html}
}

@misc{ren2025deepseekproverv2advancingformalmathematical,
      title={DeepSeek-Prover-V2: Advancing Formal Mathematical Reasoning via Reinforcement Learning for Subgoal Decomposition}, 
      author={Z. Z. Ren and Zhihong Shao and Junxiao Song and Huajian Xin and Haocheng Wang and Wanjia Zhao and Liyue Zhang and Zhe Fu and Qihao Zhu and Dejian Yang and Z. F. Wu and Zhibin Gou and Shirong Ma and Hongxuan Tang and Yuxuan Liu and Wenjun Gao and Daya Guo and Chong Ruan},
      year={2025},
      eprint={2504.21801},
      archivePrefix={arXiv},
      primaryClass={cs.CL},
      url={https://arxiv.org/abs/2504.21801}, 
}

@inproceedings{10.5555/3294996.3295057,
author = {Namkoong, Hongseok and Duchi, John C.},
title = {Variance-based regularization with convex objectives},
year = {2017},
isbn = {9781510860964},
publisher = {Curran Associates Inc.},
address = {Red Hook, NY, USA},
booktitle = {Proceedings of the 31st International Conference on Neural Information Processing Systems},
pages = {2975–2984},
numpages = {10},
location = {Long Beach, California, USA},
series = {NIPS'17}
}

@article{DBLP:journals/corr/Shalev-ShwartzW16,
  author       = {Shai Shalev{-}Shwartz and
                  Yonatan Wexler},
  title        = {Minimizing the Maximal Loss: How and Why?},
  journal      = {CoRR},
  volume       = {abs/1602.01690},
  year         = {2016},
  url          = {http://arxiv.org/abs/1602.01690},
  eprinttype    = {arXiv},
  eprint       = {1602.01690},
  bibsource    = {dblp computer science bibliography, https://dblp.org}
}

@inproceedings{yuan2021large,
  title={Large-scale robust deep auc maximization: A new surrogate loss and empirical studies on medical image classification},
  author={Yuan, Zhuoning and Yan, Yan and Sonka, Milan and Yang, Tianbao},
  booktitle={Proceedings of the IEEE/CVF International Conference on Computer Vision},
  pages={3040--3049},
  year={2021}
}

@inproceedings{burges2005learning,
  title={Learning to rank using gradient descent},
  author={Burges, Chris and Shaked, Tal and Renshaw, Erin and Lazier, Ari and Deeds, Matt and Hamilton, Nicole and Hullender, Greg},
  booktitle={Proceedings of the 22nd international conference on Machine learning},
  pages={89--96},
  year={2005}
}

@article{freund2003efficient,
  title={An efficient boosting algorithm for combining preferences},
  author={Freund, Yoav and Iyer, Raj and Schapire, Robert E and Singer, Yoram},
  journal={Journal of machine learning research},
  volume={4},
  number={Nov},
  pages={933--969},
  year={2003}
}

@misc{bercovich2025llamanemotronefficientreasoningmodels,
      title={Llama-Nemotron: Efficient Reasoning Models}, 
      author={Akhiad Bercovich and Itay Levy and Izik Golan and Mohammad Dabbah and Ran El-Yaniv and Omri Puny and Ido Galil and Zach Moshe and Tomer Ronen and Najeeb Nabwani and Ido Shahaf and Oren Tropp and Ehud Karpas and Ran Zilberstein and Jiaqi Zeng and Soumye Singhal and Alexander Bukharin and Yian Zhang and Tugrul Konuk and Gerald Shen and Ameya Sunil Mahabaleshwarkar and Bilal Kartal and Yoshi Suhara and Olivier Delalleau and Zijia Chen and Zhilin Wang and David Mosallanezhad and Adi Renduchintala and Haifeng Qian and Dima Rekesh and Fei Jia and Somshubra Majumdar and Vahid Noroozi and Wasi Uddin Ahmad and Sean Narenthiran and Aleksander Ficek and Mehrzad Samadi and Jocelyn Huang and Siddhartha Jain and Igor Gitman and Ivan Moshkov and Wei Du and Shubham Toshniwal and George Armstrong and Branislav Kisacanin and Matvei Novikov and Daria Gitman and Evelina Bakhturina and Jane Polak Scowcroft and John Kamalu and Dan Su and Kezhi Kong and Markus Kliegl and Rabeeh Karimi and Ying Lin and Sanjeev Satheesh and Jupinder Parmar and Pritam Gundecha and Brandon Norick and Joseph Jennings and Shrimai Prabhumoye and Syeda Nahida Akter and Mostofa Patwary and Abhinav Khattar and Deepak Narayanan and Roger Waleffe and Jimmy Zhang and Bor-Yiing Su and Guyue Huang and Terry Kong and Parth Chadha and Sahil Jain and Christine Harvey and Elad Segal and Jining Huang and Sergey Kashirsky and Robert McQueen and Izzy Putterman and George Lam and Arun Venkatesan and Sherry Wu and Vinh Nguyen and Manoj Kilaru and Andrew Wang and Anna Warno and Abhilash Somasamudramath and Sandip Bhaskar and Maka Dong and Nave Assaf and Shahar Mor and Omer Ullman Argov and Scot Junkin and Oleksandr Romanenko and Pedro Larroy and Monika Katariya and Marco Rovinelli and Viji Balas and Nicholas Edelman and Anahita Bhiwandiwalla and Muthu Subramaniam and Smita Ithape and Karthik Ramamoorthy and Yuting Wu and Suguna Varshini Velury and Omri Almog and Joyjit Daw and Denys Fridman and Erick Galinkin and Michael Evans and Shaona Ghosh and Katherine Luna and Leon Derczynski and Nikki Pope and Eileen Long and Seth Schneider and Guillermo Siman and Tomasz Grzegorzek and Pablo Ribalta and Monika Katariya and Chris Alexiuk and Joey Conway and Trisha Saar and Ann Guan and Krzysztof Pawelec and Shyamala Prayaga and Oleksii Kuchaiev and Boris Ginsburg and Oluwatobi Olabiyi and Kari Briski and Jonathan Cohen and Bryan Catanzaro and Jonah Alben and Yonatan Geifman and Eric Chung},
      year={2025},
      eprint={2505.00949},
      archivePrefix={arXiv},
      primaryClass={cs.CL},
      url={https://arxiv.org/abs/2505.00949}, 
}

@misc{xue2025dancegrpounleashinggrpovisual,
      title={DanceGRPO: Unleashing GRPO on Visual Generation}, 
      author={Zeyue Xue and Jie Wu and Yu Gao and Fangyuan Kong and Lingting Zhu and Mengzhao Chen and Zhiheng Liu and Wei Liu and Qiushan Guo and Weilin Huang and Ping Luo},
      year={2025},
      eprint={2505.07818},
      archivePrefix={arXiv},
      primaryClass={cs.CV},
      url={https://arxiv.org/abs/2505.07818}, 
}

@misc{lambert2025tulu3pushingfrontiers,
      title={Tulu 3: Pushing Frontiers in Open Language Model Post-Training}, 
      author={Nathan Lambert and Jacob Morrison and Valentina Pyatkin and Shengyi Huang and Hamish Ivison and Faeze Brahman and Lester James V. Miranda and Alisa Liu and Nouha Dziri and Shane Lyu and Yuling Gu and Saumya Malik and Victoria Graf and Jena D. Hwang and Jiangjiang Yang and Ronan Le Bras and Oyvind Tafjord and Chris Wilhelm and Luca Soldaini and Noah A. Smith and Yizhong Wang and Pradeep Dasigi and Hannaneh Hajishirzi},
      year={2025},
      eprint={2411.15124},
      archivePrefix={arXiv},
      primaryClass={cs.CL},
      url={https://arxiv.org/abs/2411.15124}, 
}

@misc{guo2025discriminativefinetuninggenerativelarge,
      title={Discriminative Finetuning of Generative Large Language Models without Reward Models and Preference Data}, 
      author={Siqi Guo and Ilgee Hong and Vicente Balmaseda and Tuo Zhao and Tianbao Yang},
      year={2025},
      eprint={2502.18679},
      archivePrefix={arXiv},
      primaryClass={cs.CL},
      url={https://arxiv.org/abs/2502.18679}, 
}

@inproceedings{listnet,
author = {Cao, Zhe and Qin, Tao and Liu, Tie-Yan and Tsai, Ming-Feng and Li, Hang},
year = {2007},
month = {06},
pages = {129-136},
title = {Learning to Rank: From Pairwise Approach to Listwise Approach},
volume = {227},
journal = {Proceedings of the 24th International Conference on Machine Learning},
doi = {10.1145/1273496.1273513}
}

@book{bishop2006pattern,
  author = {Bishop, C.M.},
  publisher = {Springer New York},
  title = {Pattern recognition and machine learning},
  url = {http://scholar.google.com/scholar.bib?q=info:jYxggZ6Ag1YJ:scholar.google.com/&output=citation&hl=en&as_sdt=0,5&as_vis=1&ct=citation&cd=0},
  volume = 4,
  year = 2006
}

@article{10.5555/944790.944813,
author = {Crammer, Koby and Singer, Yoram},
title = {On the algorithmic implementation of multiclass kernel-based vector machines},
year = {2002},
issue_date = {3/1/2002},
publisher = {JMLR.org},
volume = {2},
issn = {1532-4435},
journal = {J. Mach. Learn. Res.},
month = mar,
pages = {265–292},
numpages = {28}
}

@article{cortes1995support,
  author = {Cortes, C. and Vapnik, V.},
  journal = {Machine Learning},
  pages = {273-297},
  title = {Support Vector Networks},
  volume = 20,
  year = 1995
}

@article{oce,
author = {Ben-Tal, Aharon and Teboulle, Marc},
year = {2007},
month = {02},
pages = {449-476},
title = {An Old-New Concept of Convex Risk Measures: The Optimized Certainty Equivalent},
volume = {17},
journal = {Mathematical Finance},
doi = {10.1111/j.1467-9965.2007.00311.x}
}

@misc{xu2025distributionallyrobustdirectpreference,
      title={Distributionally Robust Direct Preference Optimization}, 
      author={Zaiyan Xu and Sushil Vemuri and Kishan Panaganti and Dileep Kalathil and Rahul Jain and Deepak Ramachandran},
      year={2025},
      eprint={2502.01930},
      archivePrefix={arXiv},
      primaryClass={cs.LG},
      url={https://arxiv.org/abs/2502.01930}, 
}

@misc{wang2025reductionsapproachrisksensitivereinforcement,
      title={A Reductions Approach to Risk-Sensitive Reinforcement Learning with Optimized Certainty Equivalents}, 
      author={Kaiwen Wang and Dawen Liang and Nathan Kallus and Wen Sun},
      year={2025},
      eprint={2403.06323},
      archivePrefix={arXiv},
      primaryClass={cs.LG},
      url={https://arxiv.org/abs/2403.06323}, 
}

@misc{mroueh2025reinforcementlearningverifiablerewards,
      title={Reinforcement Learning with Verifiable Rewards: GRPO's Effective Loss, Dynamics, and Success Amplification}, 
      author={Youssef Mroueh},
      year={2025},
      eprint={2503.06639},
      archivePrefix={arXiv},
      primaryClass={cs.LG},
      url={https://arxiv.org/abs/2503.06639}, 
}

@article{yang2022auc,
  title={AUC maximization in the era of big data and AI: A survey},
  author={Yang, Tianbao and Ying, Yiming},
  journal={ACM computing surveys},
  volume={55},
  number={8},
  pages={1--37},
  year={2022},
  publisher={ACM New York, NY}
}

@article{yu2025dapo,
  title={Dapo: An open-source llm reinforcement learning system at scale},
  author={Yu, Qiying and Zhang, Zheng and Zhu, Ruofei and Yuan, Yufeng and Zuo, Xiaochen and Yue, Yu and Fan, Tiantian and Liu, Gaohong and Liu, Lingjun and Liu, Xin and others},
  journal={arXiv preprint arXiv:2503.14476},
  year={2025}
}

@article{schulman2017proximal,
  title={Proximal policy optimization algorithms},
  author={Schulman, John and Wolski, Filip and Dhariwal, Prafulla and Radford, Alec and Klimov, Oleg},
  journal={arXiv preprint arXiv:1707.06347},
  year={2017}
}

@inproceedings{schulman2015trust,
  title={Trust region policy optimization},
  author={Schulman, John and Levine, Sergey and Abbeel, Pieter and Jordan, Michael and Moritz, Philipp},
  booktitle={International conference on machine learning},
  pages={1889--1897},
  year={2015},
  organization={PMLR}
}

@article{li2024model,
  title={Model Developmental Safety: A Safety-Centric Method and Applications in Vision-Language Models},
  author={Li, Gang and Yu, Wendi and Yao, Yao and Tong, Wei and Liang, Yingbin and Lin, Qihang and Yang, Tianbao},
  journal={arXiv preprint arXiv:2410.03955},
  year={2024}
}

@article{shao2024deepseekmath,
  title={Deepseekmath: Pushing the limits of mathematical reasoning in open language models},
  author={Shao, Zhihong and Wang, Peiyi and Zhu, Qihao and Xu, Runxin and Song, Junxiao and Bi, Xiao and Zhang, Haowei and Zhang, Mingchuan and Li, YK and Wu, Y and others},
  journal={arXiv preprint arXiv:2402.03300},
  year={2024}
}

@article{guo2025deepseek,
  title={Deepseek-r1: Incentivizing reasoning capability in llms via reinforcement learning},
  author={Guo, Daya and Yang, Dejian and Zhang, Haowei and Song, Junxiao and Zhang, Ruoyu and Xu, Runxin and Zhu, Qihao and Ma, Shirong and Wang, Peiyi and Bi, Xiao and others},
  journal={arXiv preprint arXiv:2501.12948},
  year={2025}
}

@article{liu2025understanding,
  title={Understanding r1-zero-like training: A critical perspective},
  author={Liu, Zichen and Chen, Changyu and Li, Wenjun and Qi, Penghui and Pang, Tianyu and Du, Chao and Lee, Wee Sun and Lin, Min},
  journal={arXiv preprint arXiv:2503.20783},
  year={2025}
}

@article{su2025trust,
  title={Trust Region Preference Approximation: A simple and stable reinforcement learning algorithm for LLM reasoning},
  author={Su, Xuerui and Xie, Shufang and Liu, Guoqing and Xia, Yingce and Luo, Renqian and Jin, Peiran and Ma, Zhiming and Wang, Yue and Wang, Zun and Liu, Yuting},
  journal={arXiv preprint arXiv:2504.04524},
  year={2025}
}

@article{chen2025empirical,
  title={An Empirical Study on Eliciting and Improving R1-like Reasoning Models},
  author={Chen, Zhipeng and Min, Yingqian and Zhang, Beichen and Chen, Jie and Jiang, Jinhao and Cheng, Daixuan and Zhao, Wayne Xin and Liu, Zheng and Miao, Xu and Lu, Yang and others},
  journal={arXiv preprint arXiv:2503.04548},
  year={2025}
}

@misc{deepscaler2025,
  title={DeepScaleR: Surpassing O1-Preview with a 1.5B Model by Scaling RL},
  author={Michael Luo and Sijun Tan and Justin Wong and Xiaoxiang Shi and William Y. Tang and Manan Roongta and Colin Cai and Jeffrey Luo and Li Erran Li and Raluca Ada Popa and Ion Stoica},
  howpublished={\url{https://pretty-radio-b75.notion.site/DeepScaleR-Surpassing-O1-Preview-with-a-1-5B-Model-by-Scaling-RL-19681902c1468005bed8ca303013a4e2}},
  note={Notion Blog},
  year={2025}
}

@article{zhang2025grpo,
  title={GRPO-LEAD: A Difficulty-Aware Reinforcement Learning Approach for Concise Mathematical Reasoning in Language Models},
  author={Zhang, Jixiao and Zuo, Chunsheng},
  journal={arXiv preprint arXiv:2504.09696},
  year={2025}
}

@article{gao2024omni,
  title={Omni-math: A universal olympiad level mathematic benchmark for large language models},
  author={Gao, Bofei and Song, Feifan and Yang, Zhe and Cai, Zefan and Miao, Yibo and Dong, Qingxiu and Li, Lei and Ma, Chenghao and Chen, Liang and Xu, Runxin and others},
  journal={arXiv preprint arXiv:2410.07985},
  year={2024}
}

@article{min2024imitate,
  title={Imitate, explore, and self-improve: A reproduction report on slow-thinking reasoning systems},
  author={Min, Yingqian and Chen, Zhipeng and Jiang, Jinhao and Chen, Jie and Deng, Jia and Hu, Yiwen and Tang, Yiru and Wang, Jiapeng and Cheng, Xiaoxue and Song, Huatong and others},
  journal={arXiv preprint arXiv:2412.09413},
  year={2024}
}

@article{chen2021evaluating,
  title={Evaluating large language models trained on code},
  author={Chen, Mark and Tworek, Jerry and Jun, Heewoo and Yuan, Qiming and Pinto, Henrique Ponde De Oliveira and Kaplan, Jared and Edwards, Harri and Burda, Yuri and Joseph, Nicholas and Brockman, Greg and others},
  journal={arXiv preprint arXiv:2107.03374},
  year={2021}
}

@article{rafailov2023direct,
  title={Direct preference optimization: Your language model is secretly a reward model},
  author={Rafailov, Rafael and Sharma, Archit and Mitchell, Eric and Manning, Christopher D and Ermon, Stefano and Finn, Chelsea},
  journal={Advances in Neural Information Processing Systems},
  volume={36},
  pages={53728--53741},
  year={2023}
}

@inproceedings{azar2024general,
  title={A general theoretical paradigm to understand learning from human preferences},
  author={Azar, Mohammad Gheshlaghi and Guo, Zhaohan Daniel and Piot, Bilal and Munos, Remi and Rowland, Mark and Valko, Michal and Calandriello, Daniele},
  booktitle={International Conference on Artificial Intelligence and Statistics},
  pages={4447--4455},
  year={2024},
  organization={PMLR}
}

@article{ethayarajh2024kto,
  title={Kto: Model alignment as prospect theoretic optimization},
  author={Ethayarajh, Kawin and Xu, Winnie and Muennighoff, Niklas and Jurafsky, Dan and Kiela, Douwe},
  journal={arXiv preprint arXiv:2402.01306},
  year={2024}
}

@article{team2025kimi,
  title={Kimi k1. 5: Scaling reinforcement learning with llms},
  author={Team, Kimi and Du, Angang and Gao, Bofei and Xing, Bowei and Jiang, Changjiu and Chen, Cheng and Li, Cheng and Xiao, Chenjun and Du, Chenzhuang and Liao, Chonghua and others},
  journal={arXiv preprint arXiv:2501.12599},
  year={2025}
}

@inproceedings{xu2024contrastive,
  title={Contrastive Preference Optimization: Pushing the Boundaries of LLM Performance in Machine Translation},
  author={Xu, Haoran and Sharaf, Amr and Chen, Yunmo and Tan, Weiting and Shen, Lingfeng and Van Durme, Benjamin and Murray, Kenton and Kim, Young Jin},
  booktitle={International Conference on Machine Learning},
  pages={55204--55224},
  year={2024},
  organization={PMLR}
}

@article{meng2024simpo,
  title={Simpo: Simple preference optimization with a reference-free reward},
  author={Meng, Yu and Xia, Mengzhou and Chen, Danqi},
  journal={Advances in Neural Information Processing Systems},
  volume={37},
  pages={124198--124235},
  year={2024}
}

@article{guo2025discriminative,
  title={Discriminative Finetuning of Generative Large Language Models without Reward Models and Preference Data},
  author={Guo, Siqi and Hong, Ilgee and Balmaseda, Vicente and Zhao, Tuo and Yang, Tianbao},
  journal={arXiv preprint arXiv:2502.18679},
  year={2025}
}

@article{sutton1999policy,
  title={Policy gradient methods for reinforcement learning with function approximation},
  author={Sutton, Richard S and McAllester, David and Singh, Satinder and Mansour, Yishay},
  journal={Advances in neural information processing systems},
  volume={12},
  year={1999}
}

@book{kullback1997information,
  title={Information theory and statistics},
  author={Kullback, Solomon},
  year={1997},
  publisher={Courier Corporation}
}

@article{levy2020large,
  title={Large-scale methods for distributionally robust optimization},
  author={Levy, Daniel and Carmon, Yair and Duchi, John C and Sidford, Aaron},
  journal={Advances in Neural Information Processing Systems},
  volume={33},
  pages={8847--8860},
  year={2020}
}

@article{duchi2019variance,
  title={Variance-based regularization with convex objectives},
  author={Duchi, John and Namkoong, Hongseok},
  journal={Journal of Machine Learning Research},
  volume={20},
  number={68},
  pages={1--55},
  year={2019}
}

@article{chu2025gpg,
  title={GPG: A Simple and Strong Reinforcement Learning Baseline for Model Reasoning},
  author={Chu, Xiangxiang and Huang, Hailang and Zhang, Xiao and Wei, Fei and Wang, Yong},
  journal={arXiv preprint arXiv:2504.02546},
  year={2025}
}

@article{lewkowycz2022solving,
  title={Solving quantitative reasoning problems with language models},
  author={Lewkowycz, Aitor and Andreassen, Anders and Dohan, David and Dyer, Ethan and Michalewski, Henryk and Ramasesh, Vinay and Slone, Ambrose and Anil, Cem and Schlag, Imanol and Gutman-Solo, Theo and others},
  journal={Advances in Neural Information Processing Systems},
  volume={35},
  pages={3843--3857},
  year={2022}
}

@inproceedings{lightman2023let,
  title={Let's verify step by step},
  author={Lightman, Hunter and Kosaraju, Vineet and Burda, Yuri and Edwards, Harrison and Baker, Bowen and Lee, Teddy and Leike, Jan and Schulman, John and Sutskever, Ilya and Cobbe, Karl},
  booktitle={The Twelfth International Conference on Learning Representations},
  year={2023}
}

@article{hendrycks2021measuring,
  title={Measuring mathematical problem solving with the math dataset},
  author={Hendrycks, Dan and Burns, Collin and Kadavath, Saurav and Arora, Akul and Basart, Steven and Tang, Eric and Song, Dawn and Steinhardt, Jacob},
  journal={arXiv preprint arXiv:2103.03874},
  year={2021}
}

@article{ouyang2022training,
  title={Training language models to follow instructions with human feedback},
  author={Ouyang, Long and Wu, Jeffrey and Jiang, Xu and Almeida, Diogo and Wainwright, Carroll and Mishkin, Pamela and Zhang, Chong and Agarwal, Sandhini and Slama, Katarina and Ray, Alex and others},
  journal={Advances in neural information processing systems},
  volume={35},
  pages={27730--27744},
  year={2022}
}

@misc{skywork-or1-2025,
  title={Skywork Open Reasoner Series},
  author = {He, Jujie and Liu, Jiacai and Liu, Chris Yuhao and Yan, Rui and Wang, Chaojie and Cheng, Peng and Zhang, Xiaoyu and Zhang, Fuxiang and Xu, Jiacheng and Shen, Wei and Li, Siyuan and Zeng, Liang and Wei, Tianwen and Cheng, Cheng and An, Bo and Liu, Yang and Zhou, Yahui},
  howpublished={\url{https://capricious-hydrogen-41c.notion.site/Skywork-Open-Reaonser-Series-1d0bc9ae823a80459b46c149e4f51680}},
  note={Notion Blog},
  year={2025}
}

@misc{openr1,
  title={Open r1: A fully open reproduction of deepseek-r1.},
  author={HuggingFace},
  howpublished={\url{https://huggingface.co/blog/open-r1}},
  note={Blog},
  year={2025}
}

@misc{openo1,
  title={Learning to reason with LLMs},
  author={OpenAI},
  howpublished={\url{https://openai.com/index/learning-to-reason-with-llms/}},
  note={Blog},
  year={2024}
}

@article{wen2025light,
  title={Light-r1: Curriculum sft, dpo and rl for long cot from scratch and beyond},
  author={Wen, Liang and Cai, Yunke and Xiao, Fenrui and He, Xin and An, Qi and Duan, Zhenyu and Du, Yimin and Liu, Junchen and Tang, Lifu and Lv, Xiaowei and others},
  journal={arXiv preprint arXiv:2503.10460},
  year={2025}
}

@misc{xiaomi2025mimo,
      title={MiMo: Unlocking the Reasoning Potential of Language Model – From Pretraining to Posttraining}, 
      author={{Xiaomi LLM-Core Team}},
      year={2025},
      primaryClass={cs.CL},
      url={https://github.com/XiaomiMiMo/MiMo}, 
}

@article{lin2025cppo,
  title={Cppo: Accelerating the training of group relative policy optimization-based reasoning models},
  author={Lin, Zhihang and Lin, Mingbao and Xie, Yuan and Ji, Rongrong},
  journal={arXiv preprint arXiv:2503.22342},
  year={2025}
}

@misc{deepcoder2025,
  title={DeepCoder: A Fully Open-Source 14B Coder at O3-mini Level},
  author={Michael Luo and Sijun Tan and Roy Huang and Ameen Patel and Alpay Ariyak and Qingyang Wu and Xiaoxiang Shi and Rachel Xin and Colin Cai and Maurice Weber and Ce Zhang and Li Erran Li and Raluca Ada Popa and Ion Stoica},
  howpublished={\url{https://pretty-radio-b75.notion.site/DeepCoder-A-Fully-Open-Source-14B-Coder-at-O3-mini-Level-1cf81902c14680b3bee5eb349a512a51}},
  note={Notion Blog},
  year={2025}
}

@article{zhang2025srpo,
  title={Srpo: A cross-domain implementation of large-scale reinforcement learning on llm},
  author={Zhang, Xiaojiang and Wang, Jinghui and Cheng, Zifei and Zhuang, Wenhao and Lin, Zheng and Zhang, Minglei and Wang, Shaojie and Cui, Yinghan and Wang, Chao and Peng, Junyi and others},
  journal={arXiv preprint arXiv:2504.14286},
  year={2025}
}

@article{hu2025open,
  title={Open-reasoner-zero: An open source approach to scaling up reinforcement learning on the base model},
  author={Hu, Jingcheng and Zhang, Yinmin and Han, Qi and Jiang, Daxin and Zhang, Xiangyu and Shum, Heung-Yeung},
  journal={arXiv preprint arXiv:2503.24290},
  year={2025}
}

@article{wei2022chain,
  title={Chain-of-thought prompting elicits reasoning in large language models},
  author={Wei, Jason and Wang, Xuezhi and Schuurmans, Dale and Bosma, Maarten and Xia, Fei and Chi, Ed and Le, Quoc V and Zhou, Denny and others},
  journal={Advances in neural information processing systems},
  volume={35},
  pages={24824--24837},
  year={2022}
}

@article{muennighoff2025s1,
  title={s1: Simple test-time scaling},
  author={Muennighoff, Niklas and Yang, Zitong and Shi, Weijia and Li, Xiang Lisa and Fei-Fei, Li and Hajishirzi, Hannaneh and Zettlemoyer, Luke and Liang, Percy and Cand{\`e}s, Emmanuel and Hashimoto, Tatsunori},
  journal={arXiv preprint arXiv:2501.19393},
  year={2025}
}

@article{zelikman2022star,
  title={Star: Bootstrapping reasoning with reasoning},
  author={Zelikman, Eric and Wu, Yuhuai and Mu, Jesse and Goodman, Noah},
  journal={Advances in Neural Information Processing Systems},
  volume={35},
  pages={15476--15488},
  year={2022}
}

@article{yao2023tree,
  title={Tree of thoughts: Deliberate problem solving with large language models},
  author={Yao, Shunyu and Yu, Dian and Zhao, Jeffrey and Shafran, Izhak and Griffiths, Tom and Cao, Yuan and Narasimhan, Karthik},
  journal={Advances in neural information processing systems},
  volume={36},
  pages={11809--11822},
  year={2023}
}

@article{feng2023alphazero,
  title={Alphazero-like tree-search can guide large language model decoding and training},
  author={Feng, Xidong and Wan, Ziyu and Wen, Muning and McAleer, Stephen Marcus and Wen, Ying and Zhang, Weinan and Wang, Jun},
  journal={arXiv preprint arXiv:2309.17179},
  year={2023}
}

@article{trinh2024solving,
  title={Solving olympiad geometry without human demonstrations},
  author={Trinh, Trieu H and Wu, Yuhuai and Le, Quoc V and He, He and Luong, Thang},
  journal={Nature},
  volume={625},
  number={7995},
  pages={476--482},
  year={2024},
  publisher={Nature Publishing Group UK London}
}

@article{xin2024deepseek,
  title={Deepseek-prover-v1. 5: Harnessing proof assistant feedback for reinforcement learning and monte-carlo tree search},
  author={Xin, Huajian and Ren, ZZ and Song, Junxiao and Shao, Zhihong and Zhao, Wanjia and Wang, Haocheng and Liu, Bo and Zhang, Liyue and Lu, Xuan and Du, Qiushi and others},
  journal={arXiv preprint arXiv:2408.08152},
  year={2024}
}

@article{williams1992simple,
  title={Simple statistical gradient-following algorithms for connectionist reinforcement learning},
  author={Williams, Ronald J},
  journal={Machine learning},
  volume={8},
  pages={229--256},
  year={1992},
  publisher={Springer}
}

@article{ziegler2019fine,
  title={Fine-tuning language models from human preferences},
  author={Ziegler, Daniel M and Stiennon, Nisan and Wu, Jeffrey and Brown, Tom B and Radford, Alec and Amodei, Dario and Christiano, Paul and Irving, Geoffrey},
  journal={arXiv preprint arXiv:1909.08593},
  year={2019}
}

@article{bai2022training,
  title={Training a helpful and harmless assistant with reinforcement learning from human feedback},
  author={Bai, Yuntao and Jones, Andy and Ndousse, Kamal and Askell, Amanda and Chen, Anna and DasSarma, Nova and Drain, Dawn and Fort, Stanislav and Ganguli, Deep and Henighan, Tom and others},
  journal={arXiv preprint arXiv:2204.05862},
  year={2022}
}

@article{grattafiori2024llama,
  title={The llama 3 herd of models},
  author={Grattafiori, Aaron and Dubey, Abhimanyu and Jauhri, Abhinav and Pandey, Abhinav and Kadian, Abhishek and Al-Dahle, Ahmad and Letman, Aiesha and Mathur, Akhil and Schelten, Alan and Vaughan, Alex and others},
  journal={arXiv preprint arXiv:2407.21783},
  year={2024}
}

@article{yang2024qwen2,
  title={Qwen2. 5 technical report},
  author={Yang, An and Yang, Baosong and Zhang, Beichen and Hui, Binyuan and Zheng, Bo and Yu, Bowen and Li, Chengyuan and Liu, Dayiheng and Huang, Fei and Wei, Haoran and others},
  journal={arXiv preprint arXiv:2412.15115},
  year={2024}
}

@article{ahmadian2024back,
  title={Back to basics: Revisiting reinforce style optimization for learning from human feedback in llms},
  author={Ahmadian, Arash and Cremer, Chris and Gall{\'e}, Matthias and Fadaee, Marzieh and Kreutzer, Julia and Pietquin, Olivier and {\"U}st{\"u}n, Ahmet and Hooker, Sara},
  journal={arXiv preprint arXiv:2402.14740},
  year={2024}
}

@article{li2023remax,
  title={Remax: A simple, effective, and efficient reinforcement learning method for aligning large language models},
  author={Li, Ziniu and Xu, Tian and Zhang, Yushun and Lin, Zhihang and Yu, Yang and Sun, Ruoyu and Luo, Zhi-Quan},
  journal={arXiv preprint arXiv:2310.10505},
  year={2023}
}

@article{hu2025reinforce,
  title={Reinforce++: A simple and efficient approach for aligning large language models},
  author={Hu, Jian},
  journal={arXiv preprint arXiv:2501.03262},
  year={2025}
}

@article{luong2024reft,
  title={Reft: Reasoning with reinforced fine-tuning},
  author={Luong, Trung Quoc and Zhang, Xinbo and Jie, Zhanming and Sun, Peng and Jin, Xiaoran and Li, Hang},
  journal={arXiv preprint arXiv:2401.08967},
  volume={3},
  year={2024}
}

@article{kazemnejad2024vineppo,
  title={Vineppo: Unlocking rl potential for llm reasoning through refined credit assignment},
  author={Kazemnejad, Amirhossein and Aghajohari, Milad and Portelance, Eva and Sordoni, Alessandro and Reddy, Siva and Courville, Aaron and Roux, Nicolas Le},
  journal={arXiv preprint arXiv:2410.01679},
  year={2024}
}

@article{xie2024monte,
  title={Monte carlo tree search boosts reasoning via iterative preference learning},
  author={Xie, Yuxi and Goyal, Anirudh and Zheng, Wenyue and Kan, Min-Yen and Lillicrap, Timothy P and Kawaguchi, Kenji and Shieh, Michael},
  journal={arXiv preprint arXiv:2405.00451},
  year={2024}
}

@article{chen2024step,
  title={Step-level value preference optimization for mathematical reasoning},
  author={Chen, Guoxin and Liao, Minpeng and Li, Chengxi and Fan, Kai},
  journal={arXiv preprint arXiv:2406.10858},
  year={2024}
}

@article{cao2024survey,
  title={Survey on large language model-enhanced reinforcement learning: Concept, taxonomy, and methods},
  author={Cao, Yuji and Zhao, Huan and Cheng, Yuheng and Shu, Ting and Chen, Yue and Liu, Guolong and Liang, Gaoqi and Zhao, Junhua and Yan, Jinyue and Li, Yun},
  journal={IEEE Transactions on Neural Networks and Learning Systems},
  year={2024},
  publisher={IEEE}
}

@article{silver2017mastering,
  title={Mastering chess and shogi by self-play with a general reinforcement learning algorithm},
  author={Silver, David and Hubert, Thomas and Schrittwieser, Julian and Antonoglou, Ioannis and Lai, Matthew and Guez, Arthur and Lanctot, Marc and Sifre, Laurent and Kumaran, Dharshan and Graepel, Thore and others},
  journal={arXiv preprint arXiv:1712.01815},
  year={2017}
}

@article{racaniere2017imagination,
  title={Imagination-augmented agents for deep reinforcement learning},
  author={Racani{\`e}re, S{\'e}bastien and Weber, Th{\'e}ophane and Reichert, David and Buesing, Lars and Guez, Arthur and Jimenez Rezende, Danilo and Puigdom{\`e}nech Badia, Adri{\`a} and Vinyals, Oriol and Heess, Nicolas and Li, Yujia and others},
  journal={Advances in neural information processing systems},
  volume={30},
  year={2017}
}

@inproceedings{nagabandi2018neural,
  title={Neural network dynamics for model-based deep reinforcement learning with model-free fine-tuning},
  author={Nagabandi, Anusha and Kahn, Gregory and Fearing, Ronald S and Levine, Sergey},
  booktitle={2018 IEEE international conference on robotics and automation (ICRA)},
  pages={7559--7566},
  year={2018},
  organization={IEEE}
}

@article{feinberg2018model,
  title={Model-based value estimation for efficient model-free reinforcement learning},
  author={Feinberg, Vladimir and Wan, Alvin and Stoica, Ion and Jordan, Michael I and Gonzalez, Joseph E and Levine, Sergey},
  journal={arXiv preprint arXiv:1803.00101},
  year={2018}
}

@inproceedings{mnih2016asynchronous,
  title={Asynchronous methods for deep reinforcement learning},
  author={Mnih, Volodymyr and Badia, Adria Puigdomenech and Mirza, Mehdi and Graves, Alex and Lillicrap, Timothy and Harley, Tim and Silver, David and Kavukcuoglu, Koray},
  booktitle={International conference on machine learning},
  pages={1928--1937},
  year={2016},
  organization={PmLR}
}

@article{lillicrap2015continuous,
  title={Continuous control with deep reinforcement learning},
  author={Lillicrap, Timothy P and Hunt, Jonathan J and Pritzel, Alexander and Heess, Nicolas and Erez, Tom and Tassa, Yuval and Silver, David and Wierstra, Daan},
  journal={arXiv preprint arXiv:1509.02971},
  year={2015}
}

@inproceedings{fujimoto2018addressing,
  title={Addressing function approximation error in actor-critic methods},
  author={Fujimoto, Scott and Hoof, Herke and Meger, David},
  booktitle={International conference on machine learning},
  pages={1587--1596},
  year={2018},
  organization={PMLR}
}

@article{he2024olympiadbench,
  title={Olympiadbench: A challenging benchmark for promoting agi with olympiad-level bilingual multimodal scientific problems},
  author={He, Chaoqun and Luo, Renjie and Bai, Yuzhuo and Hu, Shengding and Thai, Zhen Leng and Shen, Junhao and Hu, Jinyi and Han, Xu and Huang, Yujie and Zhang, Yuxiang and others},
  journal={arXiv preprint arXiv:2402.14008},
  year={2024}
}
\bibliographystyle{plain}

\newpage
\appendix

\appendix
\section{More Experimental Results}
For all the experiments on 1.5B models, each run consumes 4*2 40G A100 GPUs and each training step takes approximately 6 minutes. For all the experiments on 7B models, each run consumes 1*8 80G H100 GPUs and each training step takes approximately 6.5 minutes. 

\subsection{Detailed comparison between DisCO and DisCO-b}
\label{sec:disco-b}

\begin{figure}[b]
  \centering
  \subfigure[L-ratio on 1.5B]
  {\includegraphics[width=.24\textwidth]{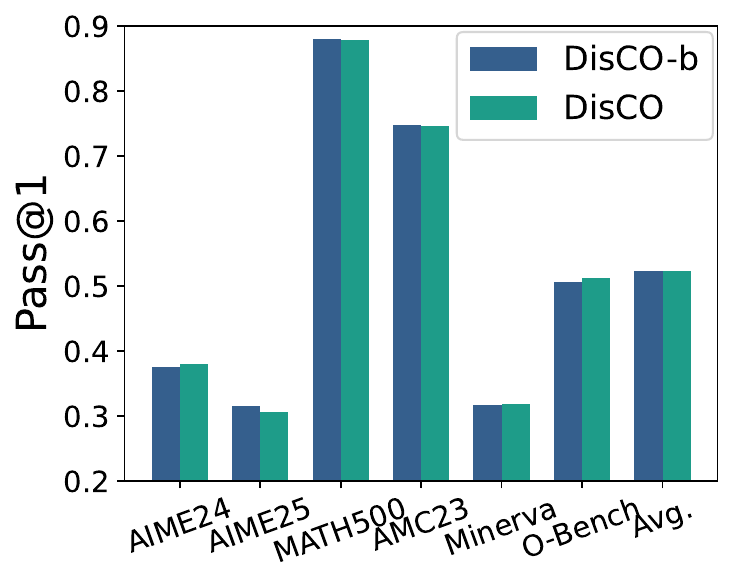}}
  \subfigure[log-L on 1.5B]
  {\includegraphics[width=.24\textwidth]{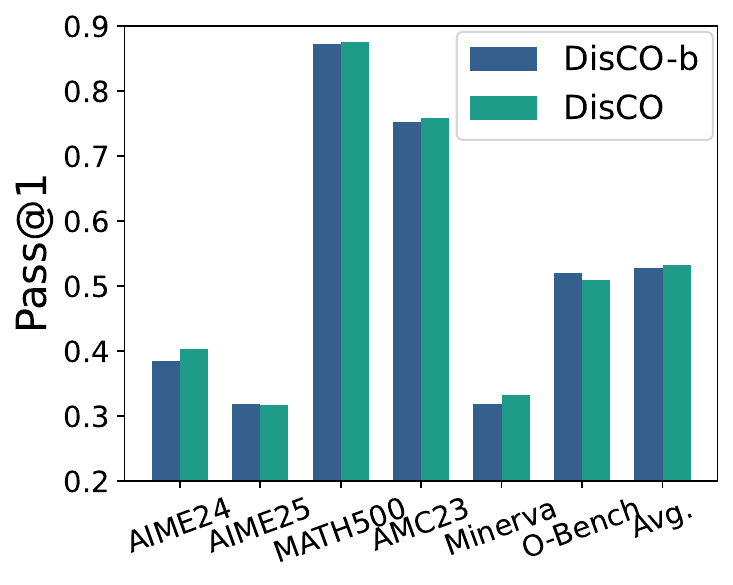}}
  \subfigure[L-ratio on 7B]
  {\includegraphics[width=.24\textwidth]{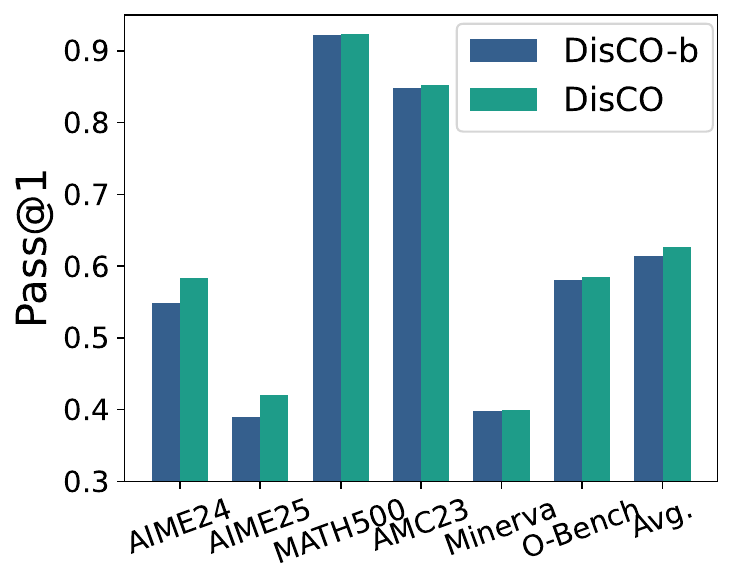}}
 \subfigure[log-L on 7B]
  {\includegraphics[width=.24\textwidth]{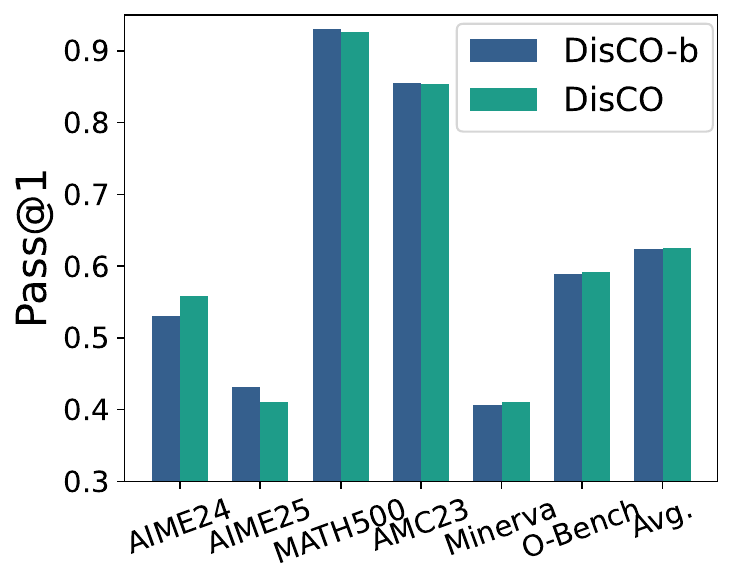}}
    \vspace*{-0.1in}
  \caption{Comparison between DisCO-b and DisCO on different models with different score functions.}
  \label{fig:disco_com}
\end{figure}

In this part, we compare DisCO and DisCO-b with different score functions on different models. As shown in Figure~\ref{fig:disco_com}, DisCO consistently demonstrates better performance compared to DisCO-b across all settings, with higher average scores observed in each case. This consistent advantage highlights the effectiveness of the full DisCO framework. Additionally, it is worth emphasizing that even the DisCO-b variants, with L-ratio or log-L scoring functions, outperform all other baseline methods that are presented in Table~\ref{tab:1.5B} and Table~\ref{tab:7B}. These results collectively underscore the robustness and general effectiveness of the DisCO approach.

\subsection{Training dynamics for fine-tuning 8B model.}
\label{sec:8b_train}

\begin{figure}[!b]
  \centering
  {\includegraphics[width=.99\textwidth]{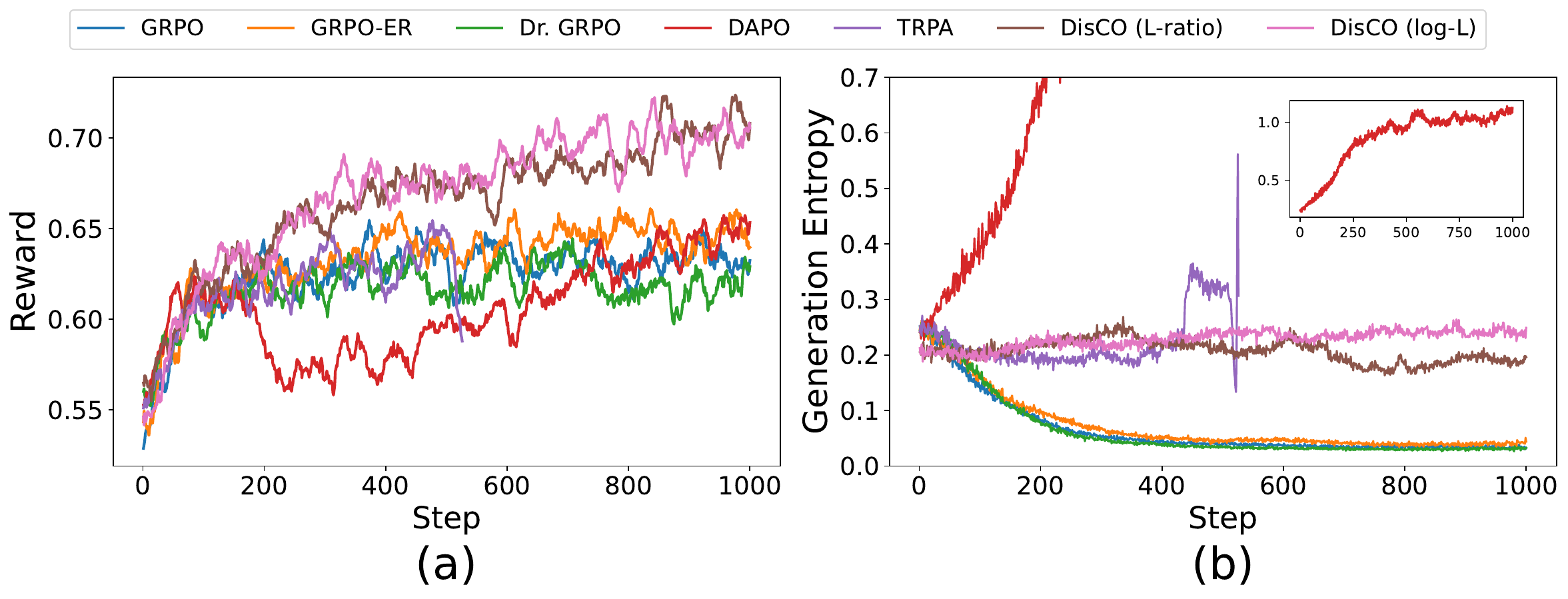}}
    \vspace*{-0.1in}
  \caption{ Training dynamics of different methods for fine-tuning  DeepSeek-R1-Distill-Llama-8B model. (a) plots the training reward (averaged over generated outputs for questions used in each step) vs the number of training steps; (b) plots the generation entropy vs training steps.}
  \label{fig:8b_train}
\end{figure}

In this part, we present the training dynamics of different methods for fine-tuning the DeepSeek-R1-Distill-Llama-8B model in Figure~\ref{fig:8b_train}. Similar to observation in Figure~\ref{fig:learning} for fine-tuning 1.5B and 7B models, we can see that GRPO, GRPO-ER, and Dr. GRPO still suffer from entropy collapse while DAPO leads to excessive entropy growth, all accompanied by premature saturation in training reward. TRPA with a KL divergence regularization is also observed with instability in the training, indicating the insufficiency of KL regularization to stabilize training. In contrast, our methods with the two scoring functions and the KL constraint demonstrate the greatest stability, with training rewards continuing to rise and generation entropy remaining around 0.2. 


\subsection{Experiments on DAPO-Math-17K dataset.}
\label{sec:dapo17k}

In order to demonstrate that the improvements achieved by DisCO are fundamental, rather than relying on specific properties of the dataset, we conducted additional experiments on the DAPO-Math-17K dataset~\cite{yu2025dapo} using 1.5B models, training them for 1400 steps. As shown in Table~\ref{tab:dapo17k}, DisCO methods still outperform other baselines by a large margin, demonstrating the generalizability of the proposed method to other datasets.

\begin{table}[t]
  \centering
  \caption{Comparison with baseline methods for fine-tuning DeepSeek-R1-Distill-Qwen-1.5B models on DAPO-Math-17K dataset.  }
  \resizebox{\linewidth}{!}{
    \begin{tabular}{lc|cccccc|c}
    \toprule
    Model/Method & MRL(Train/Test) & AIME 2024 & AIME 2025 & MATH 500 & AMC 2023 & Minerva & O-Bench & Avg. \\
    \midrule
    \cellcolor[rgb]{ .851,  .851,  .851} DS-Distill-Qwen-1.5B & 32k+   /   32k &  \cellcolor[rgb]{ .851,  .851,  .851} 0.288 & \cellcolor[rgb]{ 1,  1,  1}0.263 &  \cellcolor[rgb]{ .851,  .851,  .851} 0.828 &  \cellcolor[rgb]{ .851,  .851,  .851} 0.629 &  \cellcolor[rgb]{ .851,  .851,  .851} 0.265 &  \cellcolor[rgb]{ .851,  .851,  .851} 0.433 & \cellcolor[rgb]{ 1,  1,  1}0.451 \\
    \cellcolor[rgb]{ .851,  .851,  .851} DS-Distill-Qwen-1.5B & 32k+   /   8k & \cellcolor[rgb]{ 1,  1,  1}0.181 & \cellcolor[rgb]{ 1,  1,  1}0.215 & \cellcolor[rgb]{ 1,  1,  1}0.758 & \cellcolor[rgb]{ 1,  1,  1}0.515 & \cellcolor[rgb]{ 1,  1,  1}0.237 & \cellcolor[rgb]{ 1,  1,  1}0.353 & \cellcolor[rgb]{ 1,  1,  1}0.376 \\
        \midrule
    GRPO  & 8k   /    8k & 0.342 & 0.256 & 0.842 & 0.672 & 0.267 & 0.458 & 0.473 \\
    GRPO-ER & 8k   /   8k & 0.290 & 0.260 & 0.852 & 0.681 & 0.287 & 0.463 & 0.472 \\
    Dr. GRPO & 8k   /   8k & 0.300 & 0.250 & 0.849 & 0.705 & 0.292 & 0.464 & 0.477 \\
    DAPO  & 8k   /   8k & 0.275 & 0.229 & 0.812 & 0.653 & 0.256 & 0.441 & 0.444 \\
    TRPA  & 8k   /   8k & 0.346 & 0.279 & 0.836 & 0.683 & 0.281 & 0.450 & 0.479 \\
    DisCO (L-ratio) & 8k   /   8k & 0.413 & 0.310 & \textbf{0.874} & \textbf{0.775} & 0.307 & 0.495 & 0.529 \\
    DisCO (log-L) & 8k   /   8k & \textbf{0.460} & \textbf{0.317} & 0.873 & \textbf{0.775} & \textbf{0.320} & \textbf{0.502} & \textbf{0.541} \\
    \bottomrule
    \end{tabular}%
    }
  \label{tab:dapo17k}%
\end{table}%

\section{More Theoretical Results}
\subsection{Proof of Proposition~\ref{prop:1}}\label{app:prop}
\begin{proof}
Since $\E_{o\sim\po(\cdot|q)}r(o|q) = p(q), \text{Var}_{o\sim\po(\cdot|q)}r(o|q) = p(q)(1-p(q))$, we have
\begin{align}\label{eqn:advantage}
    A(o|q) = 
    \begin{cases}
        \sqrt{\frac{1-p(q)}{p(q)}}, & \text{ if }r(o|q)=1, \\
        -\sqrt{\frac{p(q)}{1-p(q)}}, & \text{ if } r(o|q)=0
    \end{cases}
\end{align}
According to~(\ref{eqn:dec}), we have
\begin{equation}\label{eqn:sim_grpo}
\begin{aligned}
    &\E_q\E_{o\sim\po(\cdot|q)}\bigg[\frac{1}{|o|}\sum_{t=1}^{|o|}f\bigg(\frac{\pi_\theta(o_t|q,o_{<t})}{\po(o_t|q,o_{<t})}, A(o|q)\bigg)\bigg] \\
    &=\E_q\bigg[p(q)\E_{o\sim\po^+(\cdot|q)} \frac{1}{|o|}\sum_{t=1}^{|o|}f\bigg(\frac{\pi_\theta(o_t|q,o_{<t})}{\po(o_t|q,o_{<t})}, A(o|q)\bigg)\\
    &\quad\quad\quad + (1-p(q))\E_{o\sim\po^-(\cdot|q)}\frac{1}{|o|}\sum_{t=1}^{|o|}f\bigg(\frac{\pi_\theta(o_t|q,o_{<t})}{\po(o_t|q,o_{<t})}, A(o|q)\bigg)\bigg]\\
    &=\E_q\bigg[p(q)\E_{o\sim\po^+(\cdot|q)} \frac{1}{|o|}\sum_{t=1}^{|o|}f\bigg(\frac{\pi_\theta(o_t|q,o_{<t})}{\po(o_t|q,o_{<t})}, \sqrt{\frac{1-p(q)}{p(q)}}\bigg)\\
    & \quad\quad\quad+ (1-p(q))\E_{o\sim\po^-(\cdot|q)}\frac{1}{|o|}\sum_{t=1}^{|o|}f\bigg(\frac{\pi_\theta(o_t|q,o_{<t})}{\po(o_t|q,o_{<t})}, -\sqrt{\frac{p(q)}{1-p(q)}}\bigg) \bigg]  \\
    &=\E_q\sqrt{p(q)(1-p(q))}\bigg[\E_{o\sim\po^+(\cdot|q)} \frac{1}{|o|}\sum_{t=1}^{|o|}f^+(\frac{\pi_\theta(o_t|q,o_{<t})}{\po(o_t|q,o_{<t})}, 1) \\
    &\quad\quad\quad\quad\quad\quad\quad\quad\quad\quad-\E_{o\sim\po^-(\cdot|q)}\frac{1}{|o|}\sum_{t=1}^{|o|}f^-(\frac{\pi_\theta(o_t|q,o_{<t})}{\po(o_t|q,o_{<t})}, 1 ) \bigg] 
\end{aligned}
\end{equation}
where the last equality is due to the assumption about $f(x, y)$. For GPRO, we have $f^+(x, 1) =  \min (x, \text{clip}(x, 1-\epsilon, 1+\epsilon)) = \min(x, 1+\epsilon)$ and $f^-(x, 1) =  \max (x, \text{clip}(x, 1-\epsilon, 1+\epsilon)) = \max(x, 1 -\epsilon)$. 
\end{proof}


\begin{algorithm}[t!]
\caption{Discriminative Constrained Optimization}
\label{alg:disco}
\begin{algorithmic}[1]
\STATE \textbf{Input:} Initial policy model $\pi_0$, reward function $r$, question set $\cD$, hyperparameter $\delta, \beta, \tau$.

\STATE Policy model $\pi_\theta = \pi_0$
\FOR{Step $=1,\cdots,T$}
    \STATE{Sample a batch of questions $\cB$ from $\cD$
    }
    \STATE{Update the old policy model $\po=\pi_\theta$
    }
    \STATE{For each question $q \in \cB$, sample $n$ responses $\{o_i\}_{i=1}^n \sim \po(\cdot|q)$ denoted by $S_q$ and partition it into $S^+_q$and $S^-_q$ based on rewards $r(o_i|q)\in \{0,1\}$} 
    \FOR{minibatch $\cB_m \in \cB$}
        \STATE Compute KL divergence estimator by \\ $\hat \D_{KL} = \frac{1}{\sum_{q\in\cB_m}\sum_{o\in S_q}|o|}\sum\limits_{q\in\cB_m}\sum\limits_{o\in S_q}\sum\limits_{t=1}^{|o|}\log \frac{\po(o_t|q,o_{<t})}{\pi_\theta(o_t|q,o_{<t})}$
        \STATE Compute gradient estimator of $\cJ_2(\theta)$ by\\
        $G_1 = \frac{1}{|\cB_m|}\sum\limits_{q\in \cB_m}\frac{1}{|S_q^+|}\sum\limits_{o\in S_q^+}\left(\nabla s_\theta(o,q)- \nabla \Big(\tau \log \sum\limits_{o'\in S_q^-} \exp(\frac{s_\theta(o', q)}{\tau})\Big)\right)$
        \STATE Compute gradient estimator of constraint by $G_2 = 2\beta[\hat \D_{KL}-\delta]_+ \nabla\hat \D_{KL}$
        \STATE Update $\pi_\theta $ with Adam-W using the gradient estimator $G = G_1 + G_2$
    \ENDFOR
\ENDFOR
\end{algorithmic}
\end{algorithm}

\subsection{Connection between discriminative objectives and surrogate objectives in RL}\label{sec:connection}

The score function L-ratio is inspired by the same principle as the surrogate objective in TRPO~\cite{schulman2015trust}. TRPO aims to maximize the following objective subject to a constraint:
\begin{equation}\label{eqn:trpo}
\begin{aligned}
     \max_\theta &\E_q\E_{o\sim\pi_{\theta_{old}}(\cdot|q)} \frac{1}{|o|}\sum_{t=1}^{|o|}\frac{\pi_\theta(o_t|q,o_{<t})}{\po(o_t|q,o_{<t})} A(o_t)  \\
     & s.t. \quad \D_\text{KL}(\po||\pi_{\theta}) \leq \delta.
\end{aligned}
\end{equation}
When we apply the advantage function~\eqref{eqn:advantage} to the objective, we have 
\begin{equation}\label{eqn:trpo_dri}
\begin{aligned}
    &\E_q\E_{o\sim\po(\cdot|q)}\bigg[\frac{1}{|o|}\sum_{t=1}^{|o|}\frac{\pi_\theta(o_t|q,o_{<t})}{\po(o_t|q,o_{<t})}A(o|q)\bigg] \\
    &=\E_q\bigg[p(q)\E_{o\sim\po^+(\cdot|q)} \frac{1}{|o|}\sum_{t=1}^{|o|}\frac{\pi_\theta(o_t|q,o_{<t})}{\po(o_t|q,o_{<t})}A(o|q)\\
    &\quad\quad\quad + (1-p(q))\E_{o\sim\po^-(\cdot|q)}\frac{1}{|o|}\sum_{t=1}^{|o|}\frac{\pi_\theta(o_t|q,o_{<t})}{\po(o_t|q,o_{<t})} A(o|q)\bigg]\\
    &=\E_q\bigg[p(q)\E_{o\sim\po^+(\cdot|q)} \frac{1}{|o|}\sum_{t=1}^{|o|}\frac{\pi_\theta(o_t|q,o_{<t})}{\po(o_t|q,o_{<t})}*\sqrt{\frac{1-p(q)}{p(q)}}\\
    & \quad\quad\quad+ (1-p(q))\E_{o\sim\po^-(\cdot|q)}\frac{1}{|o|}\sum_{t=1}^{|o|}\frac{\pi_\theta(o_t|q,o_{<t})}{\po(o_t|q,o_{<t})}*(-\sqrt{\frac{p(q)}{1-p(q)}}) \bigg]  \\
    &=\E_q\sqrt{p(q)(1-p(q))}\bigg[\E_{o\sim\po^+(\cdot|q)} \frac{1}{|o|}\sum_{t=1}^{|o|}\frac{\pi_\theta(o_t|q,o_{<t})}{\po(o_t|q,o_{<t})} 
    -\E_{o\sim\po^-(\cdot|q)}\frac{1}{|o|}\sum_{t=1}^{|o|}\frac{\pi_\theta(o_t|q,o_{<t})}{\po(o_t|q,o_{<t})} \bigg] 
\end{aligned}
\end{equation}
This gives the exact scoring function L-ratio: $ s_\theta(o, q) = \frac{1}{|o|}\sum_{t=1}^{|o|}\frac{\pi_{\theta}(o_t|q, o_{<t})}{\po(o_t|q, o_{<t})}$. After removing the improper weight $\sqrt{p(q)(1-p(q))}$, Eqn.~\eqref{eqn:trpo_dri} is same as Eqn.~\eqref{eqn:diso_c} with $\ell(s) = s$.

In the reinforcement learning literature, vanilla policy gradient methods also gain significant attention due to their simplicity and remarkable performance. The vanilla policy gradient(VPG) methods work by computing an estimator of the policy gradient and plugging it into a stochastic gradient algorithm. The most commonly used  surrogate objective function for gradient estimator has the form:
\begin{equation}\label{eqn:pg}
\begin{aligned}
     \mathcal{J}_{\text{VPG}}= \E_q\E_{o\sim\pi_{\theta_{old}}(\cdot|q)} \frac{1}{|o|}\sum_{t=1}^{|o|}\log \pi_\theta(o_t|q,o_{<t}) A(o_t)
\end{aligned}
\end{equation}
Similar to the derivation above, by plugging in the advantage estimator Eqn.~\eqref{eqn:advantage}, we have:
\begin{equation}\label{eqn:pg_dri}
    \begin{aligned}
        \mathcal{J}_{\text{VPG}}=\E_q\sqrt{p(q)(1-p(q))}\bigg[&\E_{o\sim\po^+(\cdot|q)} \frac{1}{|o|}\sum_{t=1}^{|o|}\log \pi_\theta(o_t|q,o_{<t})\\
    &-\E_{o\sim\po^-(\cdot|q)}\frac{1}{|o|}\sum_{t=1}^{|o|}\log \pi_\theta(o_t|q,o_{<t})\bigg] 
    \end{aligned}
\end{equation}
This directly motivate the score function log-L: $s_\theta(o, q) = \frac{1}{|o|}\sum_{t=1}^{|o|}\log \pi_\theta(o_t|q,o_{<t})$. After removing the inappropriate weight on questions, Eqn.~\eqref{eqn:pg_dri} is same as Eqn.~\eqref{eqn:diso_c} with $\ell(s) = s$.

\begin{table}[tb]
    \centering
      \caption{Weighted discriminative objectives and their scoring functions $s^+(o, q)$ and $s^-(o,q)$ for different methods, where $\sigma(\cdot)$ is the sigmoid function.}
    \resizebox{0.98\linewidth}{!}{
    \begin{tabular}{c|c}
    \toprule
         Objective &  $\E_q\omega(q)\E_{o\sim\po^+(\cdot|q), o'\sim\po^-(\cdot|q)}\ell\big(s_\theta^+(o, q)-s_\theta^-(o', q)\big)$ \\
    \midrule
         GRPO &  $\omega(q) = \sqrt{p(q)(1-p(q))}, \quad \ell(s) = s$ \\
        &  $s_\theta^+(o, q) =\frac{1}{|o|}\sum_{t=1}^{|o|}\min(\frac{\pi_\theta(o_t|q,o_{<t})}{\po(o_t|q,o_{<t})}, 1+\epsilon),\quad s_\theta^-(o', q) = \frac{1}{|o'|}\sum_{t=1}^{|o|}\max(\frac{\pi_\theta(o'_t|q,o'_{<t})}{\po(o'_t|q,o'_{<t})}, 1-\epsilon )$ \\
    \hline
         Dr. GRPO &  $\omega(q) = {p(q)(1-p(q)}, \quad \ell(s) = s$  \\
        & $s_\theta^+(o, q) =\sum_{t=1}^{|o|}\min(\frac{\pi_\theta(o_t|q,o_{<t})}{\po(o_t|q,o_{<t})}, 1+\epsilon),\quad s_\theta^-(o', q) = \sum_{t=1}^{|o'|}\max(\frac{\pi_\theta(o'_t|q,o'_{<t})}{\po(o'_t|q,o'_{<t})}, 1-\epsilon )$ \\
        
    \hline
         DAPO &  $\omega(q) = \sqrt{p(q)(1-p(q))}, \quad \ell(s) = s$ \\
        & $s_\theta^+(o, q) =\frac{1}{\E_{o\sim\po(\cdot|q)}|o|}\sum_{t=1}^{|o|}\min(\frac{\pi_\theta(o_t|q,o_{<t})}{\po(o_t|q,o_{<t})}, 1+{\epsilon_{high}}),\quad s_\theta^-(o', q) = \frac{1}{\E_{o\sim\po(\cdot|q)}|o|}\sum_{t=1}^{|o'|}\max(\frac{\pi_\theta(o'_t|q,o'_{<t})}{\po(o'_t|q,o'_{<t})}, {(1-\epsilon_{low})} )$ \\
    \hline
         GPG &  $\omega(q) = {\alpha p(q)(1-p(q)}, \quad \ell(s) = s$\\
        & $s^+_\theta(o, q) = \frac{1}{\E_{o\sim\po(\cdot|q)}|o|}\sum_{t=1}^{|o|}\log \pi_\theta(o_t|q,o_{<t}), \quad s^-_\theta(o', q)= \frac{1}{\E_{o\sim\po(\cdot|q)}|o|}\sum_{t=1}^{|o'|}\log \pi_\theta(o'_t|q,o'_{<t})$ \\
    \hline
         TRPA &  $\omega(q)=1, \quad \ell(s) = \log(\sigma(\beta(o)s))$  \\
         &  $s^+_\theta(o, q) =\sum_{t=1}^{|o|}\log\frac{\pi_\theta(o_t|q,o_{,t})}{\pi_{ref}(o_t|q,o_{<t})}, \quad s^-_\theta(o', q)=\sum_{t=1}^{|o'|}\log\frac{\pi_\theta(o'_t|q,o'_{<t})}{\pi_{ref}(o'_t|q, o'_{<t})} $\\
    \bottomrule
    \end{tabular}
    }

    \label{tab:var}
\end{table}
\subsection{Analysis of other variants of GRPO}\label{sec:aog}
In this part, we show that the other variants of GRPO still have difficulty bias on questions. 

Let's start with Dr. GRPO. In Dr. GRPO, the un-normalized advantage function is employed:
\begin{align}\label{eqn:adv_drgrpo}
    \hat A(o|q) = 
    \begin{cases}
        1-p(q), & \text{ if }r(o|q)=1, \\
        -p(q), & \text{ if } r(o|q)=0
    \end{cases}
\end{align}
With $f(x,y) = \min (xy, \text{clip}(x, 1-\epsilon, 1+\epsilon)y)$, we have 
\begin{equation}\label{eqn:drgrpo_dri}
\begin{aligned}
\cJ_{\text{Dr.GRPO}}=&\E_q\E_{o\sim\po(\cdot|q)}\bigg[\sum_{t=1}^{|o|}f\bigg(\frac{\pi_\theta(o_t|q,o_{<t})}{\po(o_t|q,o_{<t})}, \hat A(o|q)\bigg)\bigg] \\
    &=\E_q\bigg[p(q)\E_{o\sim\po^+(\cdot|q)} \sum_{t=1}^{|o|}f\bigg(\frac{\pi_\theta(o_t|q,o_{<t})}{\po(o_t|q,o_{<t})}, \hat A(o|q)\bigg)\\
    &\quad\quad\quad + (1-p(q))\E_{o\sim\po^-(\cdot|q)}\sum_{t=1}^{|o|}f\bigg(\frac{\pi_\theta(o_t|q,o_{<t})}{\po(o_t|q,o_{<t})}, \hat A(o|q)\bigg)\bigg]\\
    &=\E_q\bigg[p(q)\E_{o\sim\po^+(\cdot|q)} \sum_{t=1}^{|o|}f\bigg(\frac{\pi_\theta(o_t|q,o_{<t})}{\po(o_t|q,o_{<t})}, 1-p(q)\bigg)\\
    & \quad\quad\quad+ (1-p(q))\E_{o\sim\po^-(\cdot|q)}\sum_{t=1}^{|o|}f\bigg(\frac{\pi_\theta(o_t|q,o_{<t})}{\po(o_t|q,o_{<t})}, -p(q)\bigg) \bigg]  \\
    &=\E_q p(q)(1-p(q))\bigg[\E_{o\sim\po^+(\cdot|q)} s_\theta^+(o, q) -\E_{o\sim\po^-(\cdot|q)}s_\theta^-(o, q) \bigg] 
\end{aligned}
\end{equation}
where $s_\theta^+(o, q) =\sum_{t=1}^{|o|}\min(\frac{\pi_\theta(o_t|q,o_{<t})}{\po(o_t|q,o_{<t})}, 1+\epsilon), s_\theta^-(o, q) = \sum_{t=1}^{|o|}\max(\frac{\pi_\theta(o_t|q,o_{<t})}{\po(o_t|q,o_{<t})}, 1-\epsilon ).$ We can see that the difficult bias $\omega(q) = p(q)(1-p(q))$ on questions persists in Dr. GRPO. 

Secondly, let's reformulate the DAPO objective to show the question-level bias. With $f(x,y) = \min (xy, \text{clip}(x, 1-\epsilon_{low}, 1+\epsilon_{high})y)$ and advantage estimator~\eqref{eqn:advantage}, the expected version of DAPO is 
\begin{equation}\label{eqn:dapo_dri}
\begin{aligned}
    \cJ_{\text{DAPO}}=&\E_q\E_{o\sim\po(\cdot|q)}\bigg[\frac{1}{\E_{o\sim\po(\cdot|q)}|o|}\sum_{t=1}^{|o|}f\bigg(\frac{\pi_\theta(o_t|q,o_{<t})}{\po(o_t|q,o_{<t})}, A(o|q)\bigg)\bigg] \\
    &=\E_q\bigg[p(q)\E_{o\sim\po^+(\cdot|q)} \frac{1}{\E_{o\sim\po(\cdot|q)}|o|}\sum_{t=1}^{|o|}f\bigg(\frac{\pi_\theta(o_t|q,o_{<t})}{\po(o_t|q,o_{<t})}, A(o|q)\bigg)\\
    &\quad\quad\quad + (1-p(q))\E_{o\sim\po^-(\cdot|q)}\frac{1}{\E_{o\sim\po(\cdot|q)}|o|}\sum_{t=1}^{|o|}f\bigg(\frac{\pi_\theta(o_t|q,o_{<t})}{\po(o_t|q,o_{<t})}, A(o|q)\bigg)\bigg]\\
    &=\E_q\bigg[p(q)\E_{o\sim\po^+(\cdot|q)} \frac{1}{\E_{o\sim\po(\cdot|q)}|o|}\sum_{t=1}^{|o|}f\bigg(\frac{\pi_\theta(o_t|q,o_{<t})}{\po(o_t|q,o_{<t})}, \sqrt{\frac{1-p(q)}{p(q)}}\bigg)\\
    & \quad\quad\quad+ (1-p(q))\E_{o\sim\po^-(\cdot|q)}\frac{1}{\E_{o\sim\po(\cdot|q)}|o|}\sum_{t=1}^{|o|}f\bigg(\frac{\pi_\theta(o_t|q,o_{<t})}{\po(o_t|q,o_{<t})}, -\sqrt{\frac{p(q)}{1-p(q)}}\bigg) \bigg]  \\
    &=\E_q \sqrt{p(q)(1-p(q))}\bigg[\E_{o\sim\po^+(\cdot|q)} s_\theta^+(o, q) -\E_{o\sim\po^-(\cdot|q)}s_\theta^-(o, q) \bigg] 
\end{aligned}
\end{equation}
where $s_\theta^+(o, q) =\frac{1}{\E_{o\sim\po(\cdot|q)}|o|}\sum_{t=1}^{|o|}\min(\frac{\pi_\theta(o_t|q,o_{<t})}{\po(o_t|q,o_{<t})}, 1+\epsilon), s_\theta^-(o, q) = \frac{1}{\E_{o\sim\po(\cdot|q)}|o|}\sum_{t=1}^{|o|}\\ \max(\frac{\pi_\theta(o_t|q,o_{<t})}{\po(o_t|q,o_{<t})}, 1-\epsilon ).$ We can see that the difficult bias $\omega(q) =\sqrt{p(q)(1-p(q))}$ is placed on questions persists in DAPO. 

Thirdly, we show the difficult bias in GPG objective. In GPG, $\alpha\hat A(o|q)$ is employed as their advantage estimator. Thus, we have
\begin{equation}\label{eqn:gpg_dri}
\begin{aligned}
    \cJ_{\text{GPG}}=&\E_q\E_{o\sim\po(\cdot|q)}\bigg[\frac{1}{\E_{o\sim\po(\cdot|q)}|o|}\sum_{t=1}^{|o|} \alpha\log \pi_\theta(o_t|q,o_{<t})\hat A(o|q)\bigg] \\
    &=\E_q\bigg[p(q)\E_{o\sim\po^+(\cdot|q)} \frac{1}{\E_{o\sim\po(\cdot|q)}|o|}\sum_{t=1}^{|o|} \alpha\log \pi_\theta(o_t|q,o_{<t})\hat A(o|q)\\
    &\quad\quad\quad + (1-p(q))\E_{o\sim\po^-(\cdot|q)}\frac{1}{\E_{o\sim\po(\cdot|q)}|o|}\sum_{t=1}^{|o|}\alpha\log \pi_\theta(o_t|q,o_{<t})\hat A(o|q)\bigg]\\
    &=\E_q\bigg[p(q)\E_{o\sim\po^+(\cdot|q)} \frac{1}{\E_{o\sim\po(\cdot|q)}|o|}\sum_{t=1}^{|o|}\alpha\log \pi_\theta(o_t|q,o_{<t}) *(1-p(q))\\
    & \quad\quad\quad+ (1-p(q))\E_{o\sim\po^-(\cdot|q)}\frac{1}{\E_{o\sim\po(\cdot|q)}|o|}\sum_{t=1}^{|o|}\alpha \log \pi_\theta(o_t|q,o_{<t})*(-p(q)) \bigg]  \\
    &=\E_q \alpha p(q)(1-p(q))\bigg[\E_{o\sim\po^+(\cdot|q)} s_\theta^+(o, q) -\E_{o\sim\po^-(\cdot|q)}s_\theta^-(o, q) \bigg] 
\end{aligned}
\end{equation}
where $s_\theta^+(o, q) =\frac{1}{\E_{o\sim\po(\cdot|q)}|o|}\sum_{t=1}^{|o|}\log \pi_\theta(o_t|q,o_{<t})$, and $s_\theta^-(o, q) = \frac{1}{\E_{o\sim\po(\cdot|q)}|o|} \sum_{t=1}^{|o|}\log\pi_\theta(o_t|q,o_{<t}).$ We can see that the difficult bias $\omega(q) = \alpha p(q)(1-p(q))$ is on questions in GPG. 

Finally, we summarize the question-level weights and their score functions for other variants of GRPO in Table~\ref{tab:var}.


\end{document}